\documentclass[conference]{IEEEtran}
\IEEEoverridecommandlockouts
\usepackage{cite}
\usepackage{amsmath,amssymb,amsfonts}
\usepackage{enumitem}
\usepackage{algorithm}
\usepackage{algorithmic}
\usepackage{graphicx}
\usepackage{caption}
\usepackage{textcomp}
\usepackage{subfig}
\usepackage{multirow}
\usepackage{multicol}
\usepackage{longtable}
\usepackage{lipsum}
\usepackage{tabularx}
\usepackage{float}
\usepackage{url}
\usepackage{dblfloatfix}
\usepackage{mathtools}
\usepackage{cuted}
\usepackage{xcolor}
\newcolumntype{L}[1]{>{\raggedright\let\newline\\\arraybackslash\hspace{0pt}}m{#1}}
\newcolumntype{C}[1]{>{\centering\let\newline\\\arraybackslash\hspace{0pt}}m{#1}}
\newcolumntype{R}[1]{>{\raggedleft\let\newline\\\arraybackslash\hspace{0pt}}m{#1}}
\def\BibTeX{{\rm B\kern-.05em{\sc i\kern-.025em b}\kern-.08em
    T\kern-.1667em\lower.7ex\hbox{E}\kern-.125emX}}
    
\begin{document}

\title{A comparative study of general fuzzy min-max neural networks for pattern classification problems\\
}

\author{
\IEEEauthorblockN{Thanh Tung Khuat}
\IEEEauthorblockA{\textit{Advanced Analytics Institute} \\ \textit{Faculty of Engineering and IT} \\
\textit{University of Technology Sydney}\\
Sydney, Australia\\
thanhtung.khuat@student.uts.edu.au}
\and
\IEEEauthorblockN{Bogdan Gabrys}
\IEEEauthorblockA{\textit{Advanced Analytics Institute} \\ \textit{Faculty of Engineering and IT} \\
\textit{University of Technology Sydney}\\
Sydney, Australia\\
bogdan.gabrys@uts.edu.au}
}

\maketitle
\thispagestyle{plain}
\pagestyle{plain}

\begin{abstract}
General fuzzy min-max (GFMM) neural network is a generalization of fuzzy neural networks formed by hyperbox fuzzy sets for classification and clustering problems. Two principle algorithms are deployed to train this type of neural network, i.e., incremental learning and agglomerative learning. This paper presents a comprehensive empirical study of performance influencing factors, advantages, and drawbacks of the general fuzzy min-max neural network on pattern classification problems. The subjects of this study include (1) the impact of maximum hyperbox size, (2) the influence of the similarity threshold and measures on the agglomerative learning algorithm, (3) the effect of data presentation order, (4) comparative performance evaluation of the GFMM with other types of fuzzy min-max neural networks and prevalent machine learning algorithms. The experimental results on benchmark datasets widely used in machine learning showed overall strong and weak points of the GFMM classifier. These outcomes also informed potential research directions for this class of machine learning algorithms in the future.
\end{abstract}


\section{Introduction} \label{intro}
Pattern classification, which belongs to the class of supervised learning, aims to discover information and knowledge under data through taking advantage of the power of learning algorithms \cite{Olivas09}. It plays a crucial role in many real-world applications ranging from medical diagnostic \cite{Burger17}, electronic devices \cite{Alibart13} to tourism \cite{Li15} and energy \cite{Jokar16}.

Multi-dimensional hyperbox fuzzy sets can be used to deal with the pattern classification problems effectively by partitioning the pattern space and assigning a class label associated with a degree of certainty for each region. Each fuzzy min-max hyperbox is represented by minimum and maximum points along with a fuzzy membership function. The membership function is employed to compute the degree-of-fit of each input sample to a given hyperbox. Meanwhile, the hyperbox is continuously adjusted during the training process to cover the input patterns.

Simpson was the first one who formulated a fuzzy min-max neural network (FMNN) using hyperbox representations and proposed the training algorithms for classification \cite{Simpson92} and clustering \cite{Simpson93} problems. Since then, many researchers have paid attention to enhancing the performance of the FMNN and addressing some of its major drawbacks. Recent surveys \cite{Sayaydeh18, Khuat19} on the FMNN have divided modified variants into two groups, i.e., fuzzy min-max networks with and without contraction process. Representatives of improved models removing the contraction procedure from the training algorithms and replacing it with particular neurons for overlapping regions among hyperboxes comprise the inclusion/exclusion fuzzy hyperbox classifier \cite{Bargiela04}, the fuzzy min-max neural network with compensatory neuron \cite{Nandedkar07}, the data-core-based FMM neural network \cite{Zhang11}, and the multi-level FMM neural network \cite{Davtalab14}. However, these methods make the structure and learning algorithms complex, and thus they are hard to expand to very large datasets. In this paper, we only focus on the first group of fuzzy min-max variants using basic expansion and contraction steps with some modifications and improvements in the learning process.

Several improved versions of FMNN in the first group consist of the enhanced fuzzy min-max neural network (EFMNN) \cite{Mohammed15}, which adds more cases for the overlap verification and contraction processes, the enhanced fuzzy min-max neural network with the K-nearest hyperbox selection rule (KNEFMNN) \cite{Mohammed17}, and the general fuzzy min-max (GFMM) neural network \cite{Gabrys00}. While different improved algorithms in the first group only handle crisp input patterns, the GFMM neural network can accept both fuzzy and crisp patterns for the input data. This characteristic supports the GFMM to manage uncertainty in the input samples explicitly. Another significant modification of the GFMM is the ability to process both classification and clustering in a single model. Therefore, the GFMM can be deployed to handle many types of real-world applications, especially problems with uncertain data and the input samples in the form of intervals.

Learning algorithms of the GFMM neural network have a number of user-defined hyper-parameters, which can have a significant impact on their performance. Hence, a comparative study which illustrates the influence of hyper-parameters on the predictive accuracy is crucial for researchers to consider the applicability of the GFMM to practical problems. In addition, the study on the influence of factors on the performance of the GFMM opens the research directions towards optimizing the parameters and hyperparameters in an automatic manner. This comparative research includes assessments of the roles of configuration parameters on the predictive results of the classifiers, clarifying the efficiency and effectiveness as well as drawbacks of the GFMM in addressing the pattern classification problems, and reviewing the classification accuracy of the GFMM model in comparison to other techniques using robust evaluation approaches. Our main contributions in this study can be summarized as follows:

\begin{itemize}
    \item A comparative study of the fuzzy min-max neural network for pattern classification problems, making clear the advantages and disadvantages of each training algorithm and identifying factors influencing the performance of the GFMM neural network. Our implementations of learning algorithms for the fuzzy min-max neural networks as well as benchmark datasets are publicly available at \url{https://github.com/UTS-AAi/comparative-gfmm}
    \item We empirically evaluate the GFMM in comparison to other types of fuzzy min-max neural networks using the hyperbox expansion/contraction mechanism in the learning process as well as popular machine learning algorithms on the benchmark datasets using robust evaluation techniques, i.e., density-preserving sampling (DPS) \cite{Budka13}, parameter tuning by the grid-search method and cross-validation, as well as statistical hypothesis tests.
\end{itemize}

The rest of this paper is organized as follows. Section \ref{fmm} describes the learning algorithms of the GFMM neural network. Several existing problems and motivations are discussed in section \ref{secprob}. Experimental results and discussions are presented in section \ref{expr}. Section \ref{direct} mentions some discussions and potential research directions to improve the effectiveness of learning algorithms for the general fuzzy min-max neural network. Section \ref{conclu} concludes the findings of this study and shows some future works.

\section{General fuzzy min-max neural network} \label{fmm}
General fuzzy min-max (GFMM) neural network was proposed by Gabrys and Bargiela \cite{Gabrys00}, which is the generalization and combination of Simpson's classification and clustering neural networks within a single training algorithm. Learning process in the GFMM neural network for the classification problems comprises the formulation and adjustment of hyperboxes in the sample space \cite{Gabrys04}. A significant improvement of the GFMM network compared to the FMNN is that its inputs are hyperboxes. This feature is very convenient for representing uncertain input data, where the values are located in the acceptable range of data. To ensure the degree of membership decreasing steadily when the input pattern moves far away from the hyperbox, Gabrys and Bargiela \cite{Gabrys00} introduced a new membership function as Eq. \ref{newmem}.
\begin{equation}
    \label{newmem}
    b_i(X) = \min \limits_{j = 1}^{n}(\min([1 - f(x_j^u - w_{ij}, \gamma_j)], [1 - f(v_{ij} - x_j^l, \gamma_j)]))
\end{equation}
where $ f(z, \gamma) = \begin{cases}
    1, \quad \mbox{if } z \cdot \gamma > 1 \\
    z \cdot \gamma, \mbox{if } 0 \leq z \cdot \gamma \leq 1 \\
    0, \quad \mbox{if } z \cdot \gamma < 0
\end{cases} $; $ \gamma = \{ \gamma_1, \ldots, \gamma_n \}$ regulates the speed of decreasing of the membership values.

Unlike the FMNN, the input layer of the GFMM contains $2n$ neurons ($ n $ is the number of dimensions of data), where first $ n $ neurons correspond to $ n $ values of the lower bounds of input data, and the others are $ n $ values of the upper bounds. The connection weights between first $ n $ input nodes and hyperboxes in the middle layer form a matrix \textbf{V} representing lower bounds of the hyperboxes. The other $ n $ input nodes are connected to the middle layer by a matrix \textbf{W} showing the upper bounds of hyperboxes. In addition to $ K $ neurons corresponding to $ K $ classes in the output layer, the GFMM neural network adds a node $ c_0 $ to which unlabelled hyperboxes in the intermediate layer connect. Each hyperbox $ B_i $ in the middle layer is connected to all class nodes within the output layer. The connection weight from hyperbox $ B_i $ to the class $ c_k $ is given by the following equation:
\begin{equation}
    u_{ik} = \begin{cases} 1, \quad \mbox{if hyperbox $ B_i $ represents the class $ c_k $ } \\
    0, \quad \mbox{otherwise}
    \end{cases}
\end{equation}
The transfer function for each class node $ c_k $ realizes a union operation of fuzzy values of all hyperboxes representing that class label, defined in Eq. \ref{classval}.
\begin{equation}
    \label{classval}
    c_k = \max \limits_{i = 1}^{m} b_i \cdot u_{ik}
\end{equation}
where $ m $ is the total number of neurons in the middle layer.

Two different learning methods have been introduced to find the connection weights of the GFMM, i.e., an incremental (online) learning \cite{Gabrys00} and an agglomerative learning \cite{Gabrys02}. 

\subsection{Incremental learning} \label{gfmm_onln} \hfill\\
Incremental learning, also known as online learning, developed by Gabrys and Bargiela \cite{Gabrys00} contains the creation and adjustment processes of hyperboxes in the sample space to cover each input pattern. Generally, the algorithm includes four steps, i.e., initialization, expansion, hyperbox overlap test, and contraction, in which the last three operations are repeated.

In the initialization stage, each hyperbox which needs to be generated is initialized with the minimum point $ V_i $ being one and the maximum point $ W_i $ being zero for each dimension. By this initialization, when an input pattern presents to the network, the minimum and maximum points are automatically adjusted identically to lower and upper bounds of the input data.

Assuming that the input pattern is in the form of $ \{ X = [X^l, X^u], c_X \} $, where $ c_X $ is the label of the input sample $ X $, $ X^l = (x_1^l, \ldots, x_n^l) $ and $ X^u = (x_1^u, \ldots, x_n^u) $ are lower and upper bounds of $ X $ respectively. When $ X $ is presented to the GFMM neural network, the algorithm finds the hyperbox $ B_i $ with the highest membership value and the same class as $ c_X $ to check two expansion conditions:

\begin{itemize}
    \item maximum allowable hyperbox size $ \theta $ as Eq. \ref{exond_gfmm}:
    \begin{equation}
        \label{exond_gfmm}
        \max(w_{ij}, x_j^u) - \min(v_{ij}, x_j^l) \leq \theta, \quad \forall j \in [1, n]
    \end{equation}
    \item class label compatibility: \\
    if $ c_X = 0 $ then adjust $ B_i $ \\
    else \\ if $ class(B_i) = \begin{cases}
        0 \rightarrow \mbox{ adjust } B_i \mbox{, assign } class(B_i) = c_X \\
        c_X \rightarrow \mbox{ adjust } B_i \\
        else \rightarrow \mbox{ find another } B_i \\
    \end{cases} $
\end{itemize}
where the adjustment procedure of $ B_i $ is given as follows:

\begin{align}
v_{ij}^{new} &= \min(v_{ij}^{old}, x_j^l) \label{adjust_new_1}\\
w_{ij}^{new} &= \max(w_{ij}^{old}, x_j^u), \quad \forall j \in [1, n]
\label{adjust_new_2}
\end{align}

If all hyperboxes representing the same class with the input pattern do not meet the expansion conditions, a new hyperbox is generated to cover the input data.

If hyperbox $ B_i $ is selected and expanded in the prior step, it would be validated the overlap with other hyperboxes $ B_k $ as follows. If the class label of $ B_i $ is equal to zero, then $ B_i $ must be checked overlapping with all existing hyperboxes; otherwise, the overlap test only occurs between $ B_i $ and hyperboxes $ B_k $ representing other class labels.

The overlap test procedure is performed dimension by dimension, and for each dimension, four overlapping conditions are verified as shown in \cite{Gabrys00}. If there exists an overlapping zone between two hyperboxes, the contraction operation is employed to eliminate the overlapping region by tuning their sizes in only one dimension with the smallest overlapping value. Four corresponding cases of the contraction process can be found in detail in \cite{Gabrys00}.

In addition to setting up a fixed value of $ \theta $ at the beginning of the learning algorithm and keeping it unchanged during the training process, another implementation using adaptive values $ \theta $ was also introduced in \cite{Gabrys00}. In this way, the algorithm starts with a large value of $ \theta $, and then this value is decreased during the presentation of training data. The value of $ \theta $ is updated after each iteration as follows:

\begin{equation*}
    \theta^{new} = \varphi \cdot \theta^{old}
\end{equation*}
where the coefficient $ \varphi~(0 \leq \varphi \leq 1) $ controls the pace of decrease of $ \theta $. The learning process stops when no training patterns are misclassified or the minimum user-defined value of $ \theta_{min} $ has been reached. This study will compare the GFMM neural network with the fixed and adaptive values of the parameter $ \theta $.

\subsection{Agglomerative learning based on full similarity matrix} \hfill\\
In the incremental learning algorithm, hyperboxes are created, expanded, and contracted whenever the input pattern comes to the network. Hence, the performance of the GFMM neural network is influenced by the data presentation order. To reduce the influence of the data presentation order on the performance of the GFMM neural network, a full similarity matrix based agglomerative learning algorithm (AGGLO-SM) was introduced in \cite{Gabrys02} using all input patterns to construct hyperboxes in a bottom-up manner.

The algorithm begins with the initialization of minimum points matrix \textbf{V} and maximum points matrix \textbf{W} to the lower bounds \textbf{$X^l$} and upper bounds \textbf{$X^u$} of all input data. A similarity matrix among hyperboxes with the same class label is then computed using one of three kinds of measures as the following for each pair of hyperboxes $ B_i $ and $ B_h $

\begin{itemize}
    \item The first similarity measure is computed based on two maximum points or two minimum points of hyperboxes. To simplify in the presentation, this measure is called ``middle distance" in this work, although the similarity measures are not distance measures: \\ $ s_{ih} = s(B_i, B_h) = \min \limits_{j = 1}^{n}(\min(1 - f(w_{hj} - w_{ij}, \gamma_j), 1 - f(v_{ij} - v_{hj}, \gamma_j))) $ \\
    It is easy to see that $ s_{ih} \neq s_{hi} $, so the similarity value of $ B_i $ and $ B_h $ can be the minimum or maximum value between $ s_{ih} $ and $ s_{hi} $. If the minimum value is used, we call ``mid-min distance" measure; otherwise, ``mid-max distance" measure is deployed.
    \item The second similarity measure is calculated using the smallest gap between two hyperboxes, called ``shortest distance" in this paper: \\
    $ \widetilde{s}_{ih} = \widetilde{s}(B_i, B_h) = \min \limits_{j = 1}^{n}(\min(1 - f(v_{hj} - w_{ij}, \gamma_j), 1 - f(v_{ij} - w_{hj}, \gamma_j))) $
    
    \item The last similarity measure is computed from the longest possible distance between two hyperboxes, called ``longest distance" in this work: \\
    $ \widehat{s}_{ih} = \widehat{s}(B_i, B_h) = \min \limits_{j = 1}^{n}(\min(1 - f(w_{hj} - v_{ij}, \gamma_j), 1 - f(w_{ij} - v_{hj}, \gamma_j))) $ \\
    It is seen that both $ \widetilde{s}_{ih} $ and $ \widehat{s}_{ih}  $ have the symmetrical property.
\end{itemize}

Based on the similarity matrix, the hyperboxes would be agglomerated sequentially by finding a pair of hyperboxes with the maximum value of the similarity measure, assuming those hyperboxes are $ B_i $ and $ B_h $. Next, four following conditions have to be satisfied:

\begin{enumerate}[label=(\alph*)]
    \item Overlap test. Hyperbox formed by aggregating $ B_i $ and $ B_h $ does not overlap with any existing hyperboxes representing other classes. If any overlapping regions occur, another pair of hyperboxes is considered.
    \item Maximum hyperbox size test: \\
    $ \max(w_{ij}, w_{hj}) - \min(v_{ij}, v_{hj}) \leq \theta, \quad \forall j \in [1, n] $
    \item The minimum similarity threshold ($ \sigma $): $ s_{ih} \geq \sigma $
    \item The class compatibility test. The hyperboxes $ B_i$ and $ B_h $ represent the same class, or one or both are unlabelled.
\end{enumerate}

If all four constraints above are satisfied, the aggregation is performed as follows:

\begin{enumerate}[label=(\alph*)]
    \item Updating the coordinates of $ B_i $ using Eqs. \ref{adjust_new_1} and \ref{adjust_new_2} so that $ B_i $ represents the aggregated hyperbox.
    \item Deleting $ B_h $ from the current set of hyperboxes and updating the similarity matrix.
\end{enumerate}

The above process is repeated until no hyperboxes can be aggregated. 

\subsection{Accelerated agglomerative learning} \hfill\\
Training time of the agglomerative algorithms based on the full similarity matrix is long because their complexity is of $ O(n^3) $ \cite{Theodoridis09}. The computational expense of the AGGLO-SM algorithm is costly, especially for massive datasets, because of computation and sorting of the similarity matrix for all pairs of hyperboxes. To decrease the training time of the agglomerative learning algorithm, Gabrys \cite{Gabrys02} proposed the second agglomerative algorithm (AGGLO-2) without using the full similarity matrix when choosing and aggregating hyperboxes.
 
 The algorithm traverses the current set of hyperboxes and chooses hyperboxes, in turn, to carry out the process of aggregation. For each hyperbox $ B_i $ chosen as the first candidate, the similarity values of $ B_i $ and remaining $ m - 1 $ hyperboxes are computed. The hyperbox $ B_h $ with the highest similarity value against $ B_i $ is selected as the second candidate. The aggregation process for hyperboxes $ B_i $ and $ B_h $ is the same as in the algorithm using the full similarity matrix. If current pair of selected hyperboxes does not meet the aggregation constraints, the hyperbox with the second highest similarity value against $ B_i $ is chosen, and the above agglomerative procedure is repeated until the agglomeration occurs, or no hyperboxes can be aggregated with the current hyperbox $ B_i $.
 
After the first iteration, there are only $ m - 2 $ hyperboxes for the next processing. The algorithm continues with the next hyperbox chosen for aggregation, and the procedure mentioned above is repeated. The training algorithm terminates when going through a whole hyperboxes set, but no aggregation operation is performed.

\section{Existing Problems and motivations}\label{secprob}
Fuzzy min-max neural networks are universal approximators, which can tackle both linear and non-linear classification problems. However, these classifiers depend on the selection of hyper-parameters, such as the maximum hyperbox size. If the hyper-parameters are set well, the trained model will achieve a good performance on unseen data. Nonetheless, this is a challenging task because of the huge searching space of parameters. This study is not to optimize the hyper-parameters in an automatic manner. Instead, we assess the impact of hyper-parameters on the performance of the models for each dataset. Based on these evaluations, we can draw conclusions related to the important role of the selection of hyper-parameters with regard to predictive accuracy of models on each training dataset. As a result, when comparing various learning algorithms, we choose the best settings in the range of potential parameters based on the performance of classifiers on validation sets, which are formed by K-fold cross-validation and the density-preserving sampling method.

\begin{figure}
    \centering
    \includegraphics[width=0.5\textwidth]{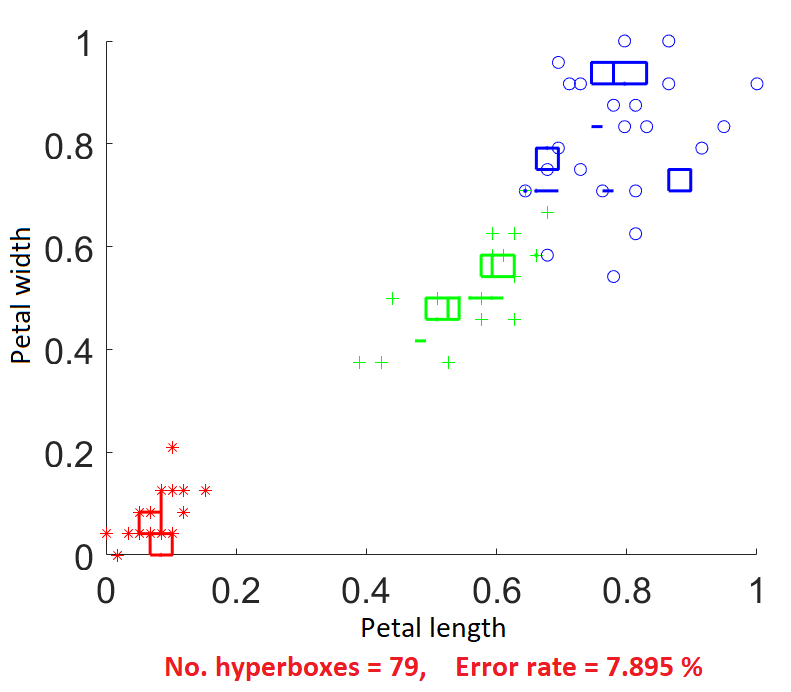}
    \caption{A hyperbox-based model is trained on the \textit{Iris} dataset}
    \label{iris_demo_full}
\end{figure}

To generate a hyperbox-based classifier with good generalization error, besides independent learning schemes such as cross-validation and resampling approaches \cite{Gabrys04}, we also need to integrate the explicit overfitting prevention mechanisms, i.e., pruning procedures, to learning algorithms. Taking decision trees as an example, if the training process constructs a full tree structure, the model will overfit the training set. Therefore, to ensure a good generalization error, one usually applies early stopping and pruning methods. Similarly, if the maximum hyperbox size is set to a small value, there are many generated hyperboxes for each hyperbox-based learner. These hyperbox fuzzy sets are more likely to overfit the training data. An example is shown in Fig. \ref{iris_demo_full} for \textit{Iris} dataset with 112 training samples and two out of its four features. The model is trained using a small value of maximum hyperbox size ($\theta = 0.06 $). It can be seen that the model contains 79 hyperboxes, and many hyperboxes include only one sample, which is unnecessarily complex.

\begin{figure}
    \centering
    \includegraphics[width=0.5\textwidth]{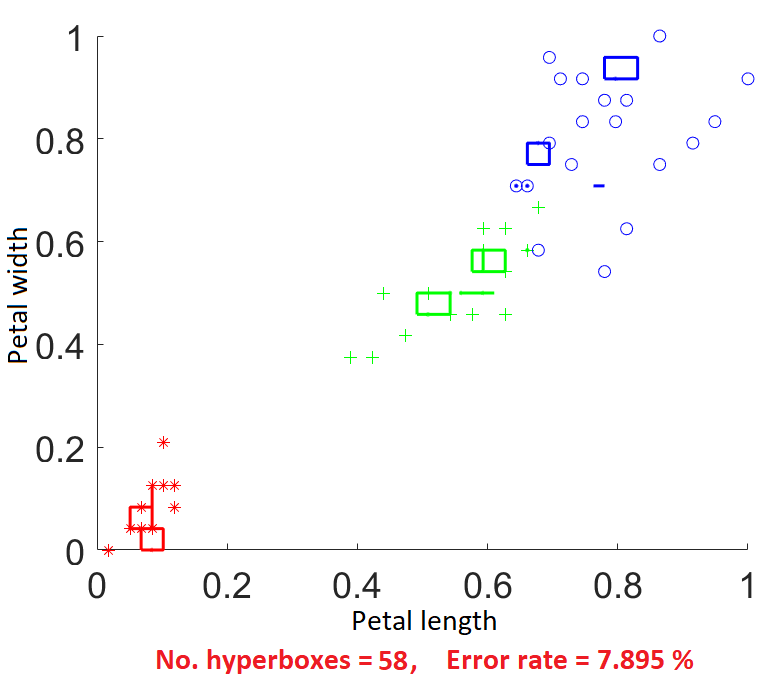}
    \caption{A hyperbox-based model is trained on the \textit{Iris} dataset}
    \label{iris_demo_part}
\end{figure}

\begin{figure}
    \centering
    \includegraphics[width=0.5\textwidth]{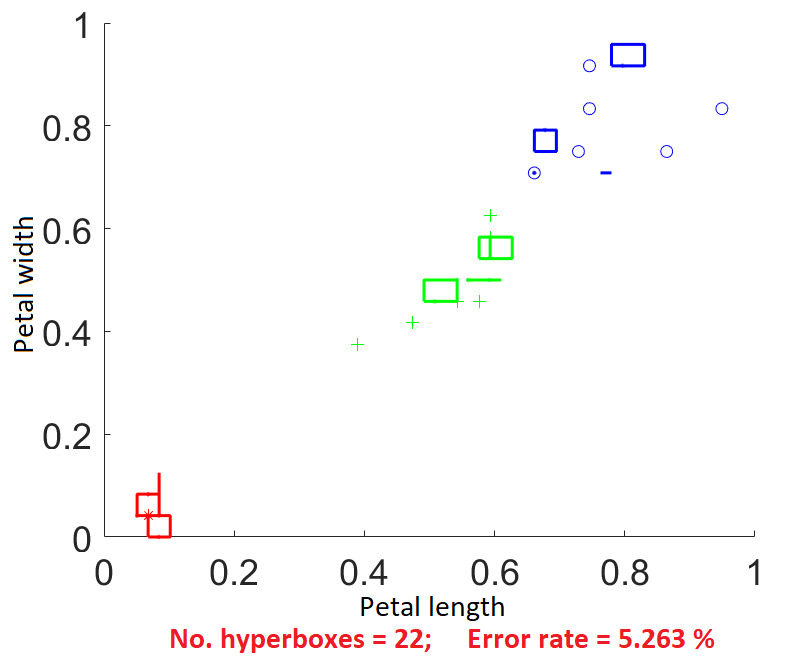}
    \caption{A hyperbox-based model is trained on the \textit{Iris} dataset}
    \label{iris_demo_pruning}
\end{figure}

To cope with this problem, we can split the training dataset into disjoint training and validation sets using the DPS method (75 training samples and 37 validation patterns). The model trained on the training set is shown in Fig. \ref{iris_demo_part}. The number of generated hyperboxes is lower than in the previous case because we used a smaller number of training samples, but the accuracy is still the same. This result also confirms that the DPS method can generate a representative training set from the original data. After training, the validation set is employed to remove low-quality hyperboxes, which have predictive accuracy less than 50\%. The final classifier is presented in Fig. \ref{iris_demo_pruning}. It can be easily observed that both the number of generated hyperboxes and error rate have been significantly reduced.

The removal of hyperboxes can lead to loss of important information because this operation is based on only the misclassification error on the validation set. If the selection of hyper-parameters results in a nearly optimal decision boundary after the training process, the pruning procedure may increase the error rates since it will break the optimal structure of the trained model. The experiments in the next section focus on clarifying the role of the pruning process if the classifier has been built using the best hyper-parameters. We also find the answer to the question of whether the impact of noisy data can be reduced through parameter settings rather than identifying and removing them through pruning or data editing \cite{Gabrys01}.

\section{Experiments and Results} \label{expr}
\subsection{Datasets}
Our experiments used 16 relatively small-sized datasets from the UCI repository \cite{Bache13}. These benchmark datasets have been widely used to evaluate machine learning algorithms such as in \cite{Salvador04}, \cite{Zelnik-Manor04}, \cite{kononenko97}, \cite{kaski03}, \cite{Tung05}, and \cite{Breiman01}. The detailed information of these datasets is shown in Table \ref{table1}. Each dataset was separated into four folds using the density-persevering sampling technique \cite{Budka13}, which is a robust and efficient method competitive to cross-validation for error estimation. Three folds were used as training data, while the remaining fold was selected as a testing set. In common, for each dataset, experiments were repeated four times with each fold used as testing data in turn and reported results were average of results on each testing fold.

\begin{table}[!ht]
\caption{Datasets were used for experiments}
\label{table1}
\centering
\begin{tabular}{|l|L{2cm}|C{1.4cm}|C{1.4cm}|C{1.3cm}|}
\hline
ID & Dataset & No. samples & No. features & No. classes \\
\hline
1 & Circle & 1000 & 3 & 2 \\ 
\hline
2 & Complex9 & 3031 & 2 & 9  \\ 
\hline
3 & Diagnostic Breast Cancer & 569 & 30 & 2  \\
\hline
4 & Glass & 214 & 9 & 6  \\
\hline
5 & Ionosphere & 351 & 34 & 2  \\
\hline
6 & Iris & 150 & 4 & 3  \\
\hline
7 & Ringnorm & 7400 & 20 & 2  \\
\hline
8 & Segmentation & 2310 & 19 & 7  \\
\hline
9 & Spherical\_5\_2 & 250 & 2 & 5  \\ 
\hline
10 & Spiral & 1000 & 2 & 2  \\
\hline
11 & Thyroid & 215 & 5 & 3  \\
\hline
12 & Twonorm & 7400 & 20 & 2  \\ 
\hline
13 & Waveform & 5000 & 21 & 3  \\ 
\hline
14 & Wine & 178 & 13 & 3  \\
\hline
15 & Yeast & 1484 & 8 & 10  \\
\hline
16 & Zelnik6 (Toy dataset) & 238 & 2 & 3  \\
\hline
\end{tabular}
\end{table}

\subsection{The influence of the maximum hyperbox size on the performance of online learning based GFMM}
\begin{figure}
    \centering
    \includegraphics[width=0.42\textwidth]{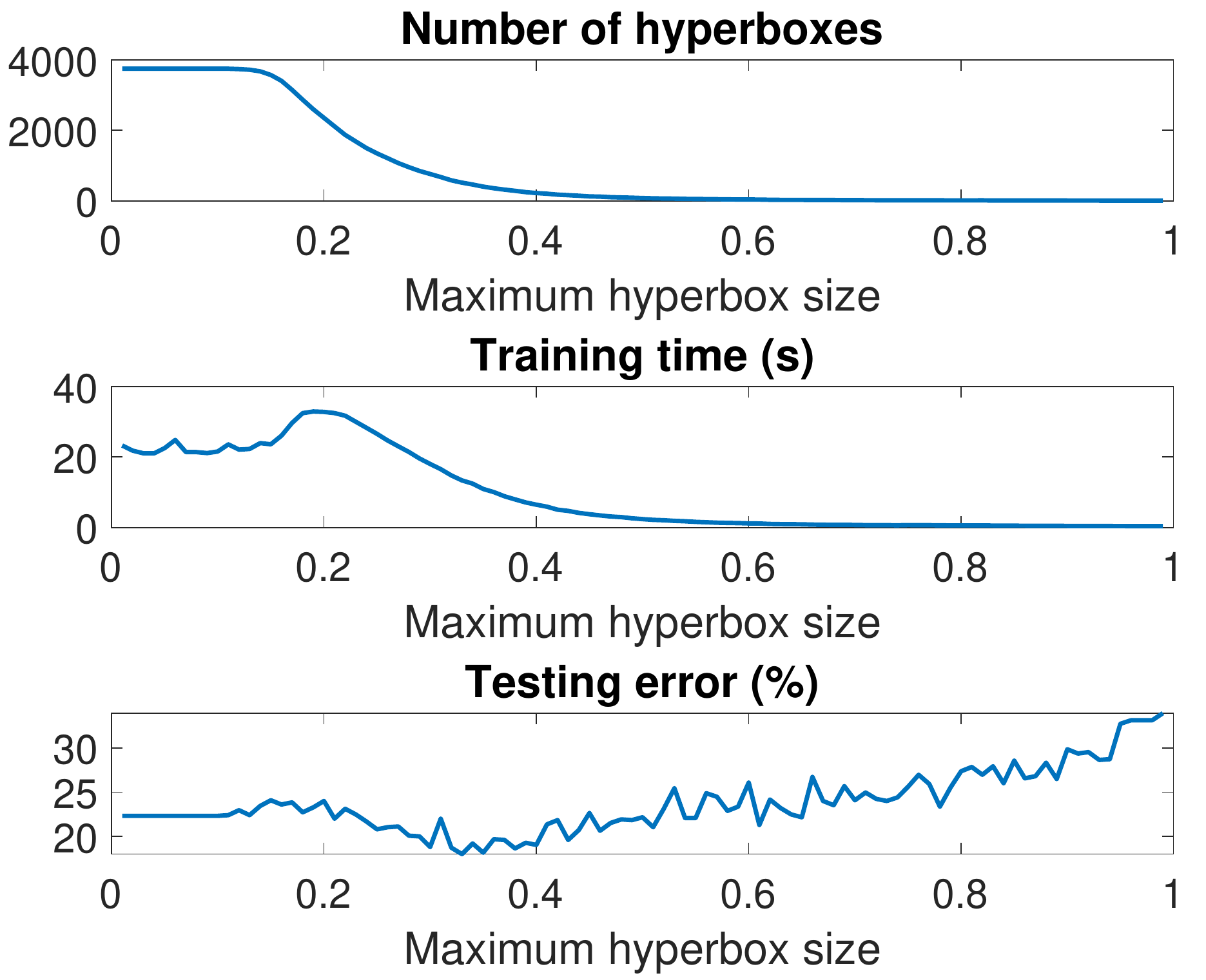}
    \caption{The change in the number of hyperboxes, training time, and testing error of the \textit{Waveform} dataset}
    \label{waveform}
\end{figure}

This experiment is to assess the impact of the maximum hyperbox size parameter, $ \theta $, on the performance of the GFMM neural network using the incremental learning algorithm. We used three out of four folds for training the network and one remaining fold for the testing process. We increased the value of $ \theta $ from 0.01 to 0.99 with the step being 0.01 and used the incremental learning with the fixed hyperbox size for each dataset. Entire figures showing the change in the number of hyperboxes, training time, and testing error of all considered datasets can be found at \url{https://github.com/UTS-AAi/comparative-gfmm/blob/master/experiment/hyperbox-size-changing.pdf}. A representative example of changing trend in the number of generated hyperboxes, training time, and testing error is presented in Fig. \ref{waveform} for the \textit{Waveform} dataset.

We can see that the larger value of $ \theta $, the fewer the number of hyperboxes in the model is generated. Generally, the training time also reduces when increasing the value of $ \theta $, and the training time is usually fast and decreases in a stable manner if the maximum hyperbox size is larger than 0.5. Furthermore, training time frequently fluctuates and stands at a high value when the value of $ \theta $ is less than 0.2. Regarding the testing error, there is no general rule for all datasets when the value of $ \theta $ gets larger, but the error rates are frequently high if the $ \theta $ thresholds are larger than 0.8, except \textit{Zelnik6}, \textit{Thyroid}, \textit{Iris}, and \textit{Wine} datasets. It is easily observed from the images that the prediction results of the GFMM using an incremental learning algorithm are substantially influenced by the selection of values of $ \theta $. It is not straightforward to choose an optimal value of $ \theta $ to gain the best performance for each dataset. Several optimization algorithms can be deployed to find the optimal value of $ \theta $ in an automatic manner.

\begin{table*} [!ht]
\centering
\caption{Comparison of fixed and adaptive maximum hyperbox size parameters ($ \theta = 0.26 $)} \label{fix_adapt_teta}
\scriptsize
{
\begin{tabular}{|p{0.02\textwidth}|p{0.14\textwidth}|p{0.1\textwidth}|p{0.11\textwidth}|p{0.11\textwidth}|p{0.1\textwidth}|p{0.11\textwidth}|p{0.11\textwidth}|}
\hline 
	\multirow{2}{*}{\textbf{ID}} & \multirow{2}{*}{\textbf{Dataset}} & \multicolumn{3}{c|}{\textbf{Fixed value}} & \multicolumn{3}{c|}{\textbf{Adaptive value ($\theta_{min} = 0.01$)}}\\
	\cline{3-8}
	& & \textbf{No. hyperboxes} & \textbf{Training time (s)} & \textbf{Testing error (\%)} & \textbf{No. hyperboxes} & \textbf{Training time (s)} & \textbf{Testing error (\%)} \\
	\hline
1  & Circle                 & 29.950    & 0.092  & 5.240       & 71.175   & 3.092 & 3.530       \\ \hline
2  & Complex 9              & 28.275   & 0.272  & 1.755     & 38.350    & 10.913  & 0.267  \\ \hline
3  & Diagnostic Breast Cancer & 113.550   & 0.302 & 4.586  & 118.400 & 0.740 & 4.516  \\ \hline
4  & Glass  & 42.675   & 0.060 & 39.286 & 75.425   & 1.220    & 40.597 \\ \hline
5  & ionosphere & 144.675  & 0.178 & 12.229  & 144.675  & 0.230  & 12.229  \\ \hline
6  & Iris & 16.775   & 0.016   & 4.683    & 18.975   & 0.393   & 4.491  \\ \hline
7  & Ringnorm   & 1411.525 & 31.666 & 26.468 & 2260.450  & 164.892 & 27.886 \\ \hline
8  & Segmentation   & 230.275  & 2.970   & 4.588   & 246.750   & 25.998  & 4.567   \\ \hline
9  & Spherical\_5\_2        & 13.600     & 0.020 & 1.274    & 13.600     & 0.040  & 1.274    \\ \hline
10 & Spiral  & 26.95    & 0.102  & 7.810  & 42.450  & 2.902 & 0.650 \\ \hline
11 & Thyroid & 22.475   & 0.025 & 4.268    & 30.400    & 0.576 & 3.988  \\ \hline
12 & Twonorm                & 1862.950  & 44.715 & 4.932    & 1926.500   & 57.923 & 4.928   \\ \hline
13 & Waveform               & 1185.700   & 24.529 & 20.688     & 1622.375 & 55.546 & 20.638     \\ \hline
14 & Wine                   & 75.375   & 0.056 & 4.229  & 75.375   & 0.074 & 4.229  \\ \hline
15 & Yeast                  & 128.900    & 0.992 & 67.832 & 1456.750  & 137.667 & 72.062   \\ \hline
16 & Zelnik6                & 12.600     & 0.015 & 0.212  & 12.600     & 0.031  & 0.212  \\ \hline
\end{tabular}
}
\end{table*}

\begin{figure}
    \centering
    \includegraphics[width=0.5\textwidth]{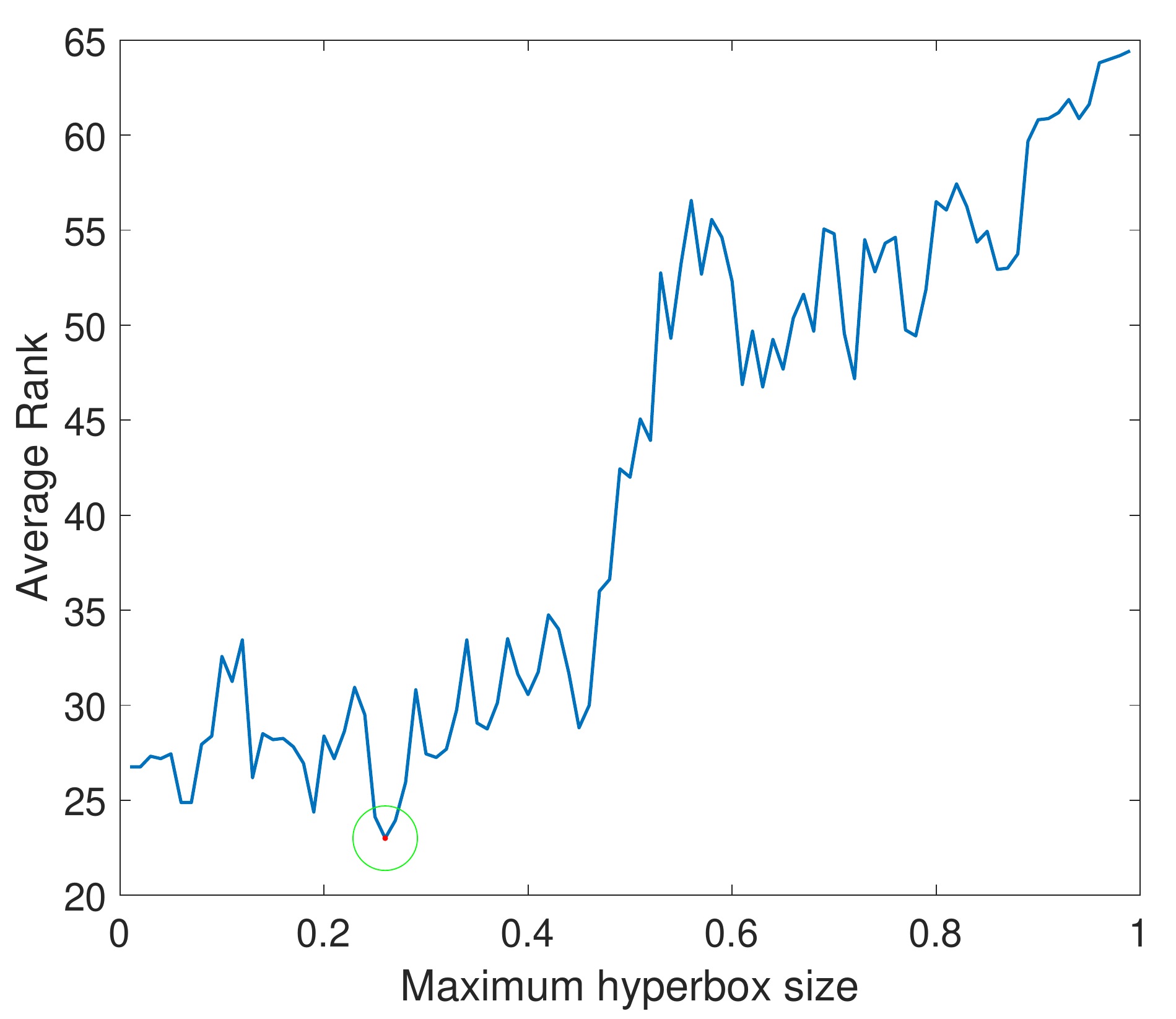}
    \caption{Average rank of performance on 16 datasets using different values of $ \theta $}
    \label{avgtheta}
\end{figure}

To remedy the impact of the maximum hyperbox size, the incremental learning algorithm using the adaptive value of $ \theta $ was developed as described at the end of subsection \ref{gfmm_onln}. To compare the performance of GFMM using the adaptive values of $ \theta $ with the one using the fixed value of $ \theta $, we selected $ \theta = 0.26 $ as the initial value, and the learning algorithm was repeated until the minimum value of $\theta $ being 0.01 was reached out ($\varphi = 0.9$) in the case of using the adaptive incremental learning algorithm. The value $ \theta = 0.26 $ was selected because it gave the lowest average rank of prediction errors over 16 datasets in comparison to other fixed values of $ \theta $ as shown in the above experiment. The average rank of the performance of general fuzzy min-max neural network using different fixed values of $ \theta $ over 16 datasets is given in Fig. \ref{avgtheta}.

In this experiment, each dataset was also split into four folds, and each execution used a fold for testing and three remaining folds were deployed for training. For each training dataset, ten runs were performed, and each iteration shuffled training data randomly. The obtained value for each testing fold is an average of ten executions. Table \ref{fix_adapt_teta} reports the averaged experimental results concerning the number of generated hyperboxes, training time, and testing error rate for two strategies of employing the value of $ \theta $ on four folds over different datasets. 

\begin{figure}
    \centering
    \includegraphics[width=0.42\textwidth]{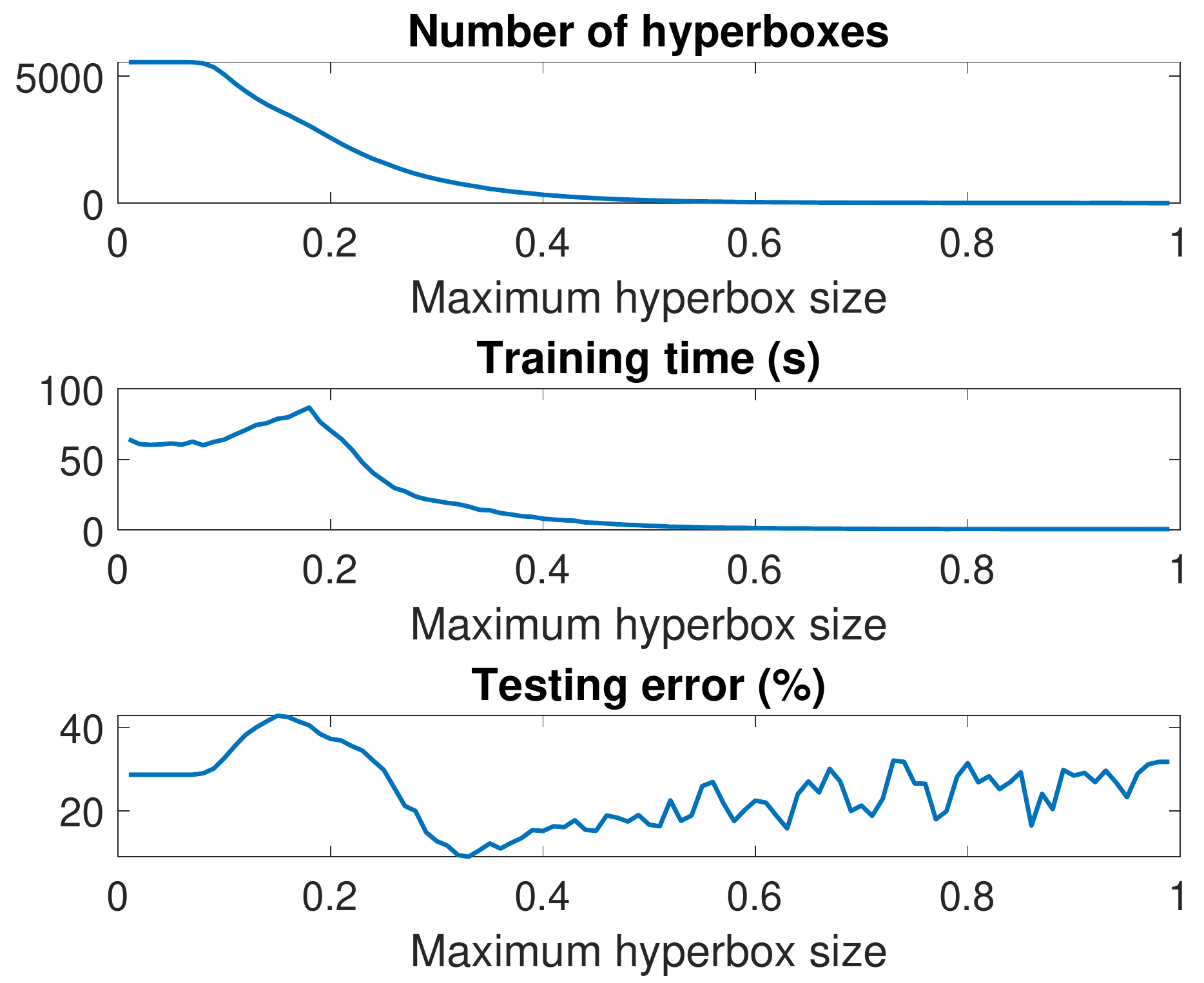}
    \caption{The change in the number of hyperboxes, training time, and testing error of the \textit{Ringnorm} dataset}
    \label{ringnorm}
\end{figure}

In several datasets such as \textit{Circle}, \textit{Complex 9}, and \textit{Spiral}, the testing errors fell sharply when using the adaptive mechanism for $ \theta $. Meanwhile, the error rate in some datasets like \textit{Glass}, \textit{Ringnorm}, and \textit{Yeast} increased slightly in the case of implementing adaptive values of $\theta$. A reason for this fact is the overfitting in the trained model. We can see this phenomenon in Fig. \ref{ringnorm} for the \textit{Ringnorm} dataset, where a large number of hyperboxes were generated and the testing errors at fixed values of $ \theta < 0.26 $ are relatively high. In the remaining cases, the error rates of the GFMM using adaptive values of $\theta$ are slightly lower or the same as those employing the fixed values of the maximum hyperbox size. We can conclude that the adaptive hyperbox size based GFMM has limited impact in case of using the starting value of $\theta$ being the best value for many datasets. To further evaluate the performance of the GFMM using the adaptive values of hyperbox size, we chose another starting value of $\theta$ away far from the optimal value. We selected $ \theta = 0.56 $ because it leads to the large changing in the average rank of GFMM as shown in Fig. \ref{avgtheta}. The outcomes of GFMM using fixed value of $ \theta = 0.56 $ and adaptive values starting from $ \theta = 0.56 $ are shown in Table \ref{adp0_56}.

\begin{table*}[!ht]
\centering
\caption{Comparison of fixed and adaptive maximum hyperbox size parameters ($ \theta = 0.56$)}
\label{adp0_56}
\scriptsize
{
\begin{tabular}{|p{0.02\textwidth}|p{0.14\textwidth}|p{0.1\textwidth}|p{0.11\textwidth}|p{0.11\textwidth}|p{0.1\textwidth}|p{0.11\textwidth}|p{0.11\textwidth}|}
\hline 
	\multirow{2}{*}{\textbf{ID}} & \multirow{2}{*}{\textbf{Dataset}} & \multicolumn{3}{c|}{\textbf{Fixed value}} & \multicolumn{3}{c|}{\textbf{Adaptive value ($\theta_{min} = 0.01$)}}\\
	\cline{3-8}
	& & \textbf{No. hyperboxes} & \textbf{Training time (s)} & \textbf{Testing error (\%)} & \textbf{No. hyperboxes} & \textbf{Training time (s)} & \textbf{Testing error (\%)} \\
	\hline
1                   & Circle                   & 9.65           & 0.059             & 15.22              & 73.275           & 3.937              & 3.48                \\ \hline
2                   & Complex 9                & 11.775         & 0.234             & 11.943             & 37.65            & 13.256             & 0.432               \\ \hline
3                   & Diagnostic Breast Cancer   & 22.35          & 0.065             & 5.733              & 84.85            & 1.571              & 4.994               \\ \hline
4                   & Glass                    & 17.225         & 0.04              & 47.983             & 105.375          & 1.793              & 46.062              \\ \hline
5                   & ionosphere               & 80.825         & 0.112             & 13.62              & 81.625           & 2.38               & 13.733              \\ \hline
6                   & Iris                     & 6.875          & 0.008             & 6.01               & 13.125           & 0.587              & 3.558               \\ \hline
7                   & Ringnorm                 & 59.25          & 1.74              & 21.77              & 2151.5           & 593.817            & 4.768               \\ \hline
8                   & Segmentation             & 47.725         & 0.486             & 17.349             & 442.9            & 34.322             & 17.882              \\ \hline
9                   & Spherical\_5\_2          & 5              & 0.014             & 0.794              & 5                & 0.032              & 0.794               \\ \hline
10                  & Spiral                   & 8.975          & 0.084             & 41.94              & 52.225           & 4.068              & 1.38                \\ \hline
11                  & Thyroid                  & 8.05           & 0.015             & 5.196              & 30.875           & 0.84               & 5.206               \\ \hline
12                  & Twonorm                  & 51.55          & 1.874             & 13.205             & 3539.95          & 561.18             & 5.27                \\ \hline
13                  & Waveform                 & 47.95          & 1.508             & 23.054             & 3265.75          & 858.192            & 19.416              \\ \hline
14                  & Wine                     & 17.7           & 0.025             & 3.586              & 17.775           & 0.037              & 3.586               \\ \hline
15                  & Yeast                    & 34.775         & 0.704             & 92.507             & 1933.275         & 626.437            & 93.713              \\ \hline
16                  & Zelnik6                  & 7              & 0.012             & 6.895              & 8.475            & 0.394              & 1.013               \\ \hline
\end{tabular}
}
\end{table*}

It is easily observed that in most of the datasets the testing errors using adaptive values of $\theta$ are significantly enhanced compared to the cases using the fixed values of $\theta$. In several datasets such as \textit{Yeast}, \textit{Thyroid}, \textit{Segmentation}, and \textit{Ionosphere}, the accuracy of predictive results decreases slightly. In general, the accuracy of GFMM using adaptive values of $\theta$ starting from $\theta = 0.56$ is superior to that employing the fixed value $\theta = 0.56$. However, the number of created hyperboxes and training time of the algorithm using the adaptive values of $ \theta $ increased considerably, especially in large-sized datasets such as \textit{Ringnorm}, \textit{Twonorm}, \textit{Waveform}, and \textit{Yeast} datasets. In addition, the accuracy of GFMM in this experiment is lower than that using adaptive values of the maximum hyperbox size starting from $\theta = 0.26$. In many datasets, it can be seen that the error rates of GFMM using the adaptive values from $ \theta = 0.56 $ are higher than those utilizing fixed value $\theta = 0.26$. These results indicate the impacts of choosing the suitable values of maximum hyperbox size on the accuracy of predictive results. They also confirm that the incremental learning algorithm using the adaptive values of the maximum hyperbox size has not yet been an effective method to tackle the dependence of classification performance on the selection of the maximum hyperbox size parameter. Hence, to compare the performance of GFMM with other methods, we will use the fixed value of $\theta$ that leads to the minimum error on the validation set in the range of given values for each dataset rather than using the same value of $\theta$ for all considered datasets.

\subsection{The influence of the similarity threshold on the performance of the agglomerative learning based GFMM using different similarity measures}
\begin{figure*}
	\centering
	\includegraphics[width=0.85\textwidth, height=4in]{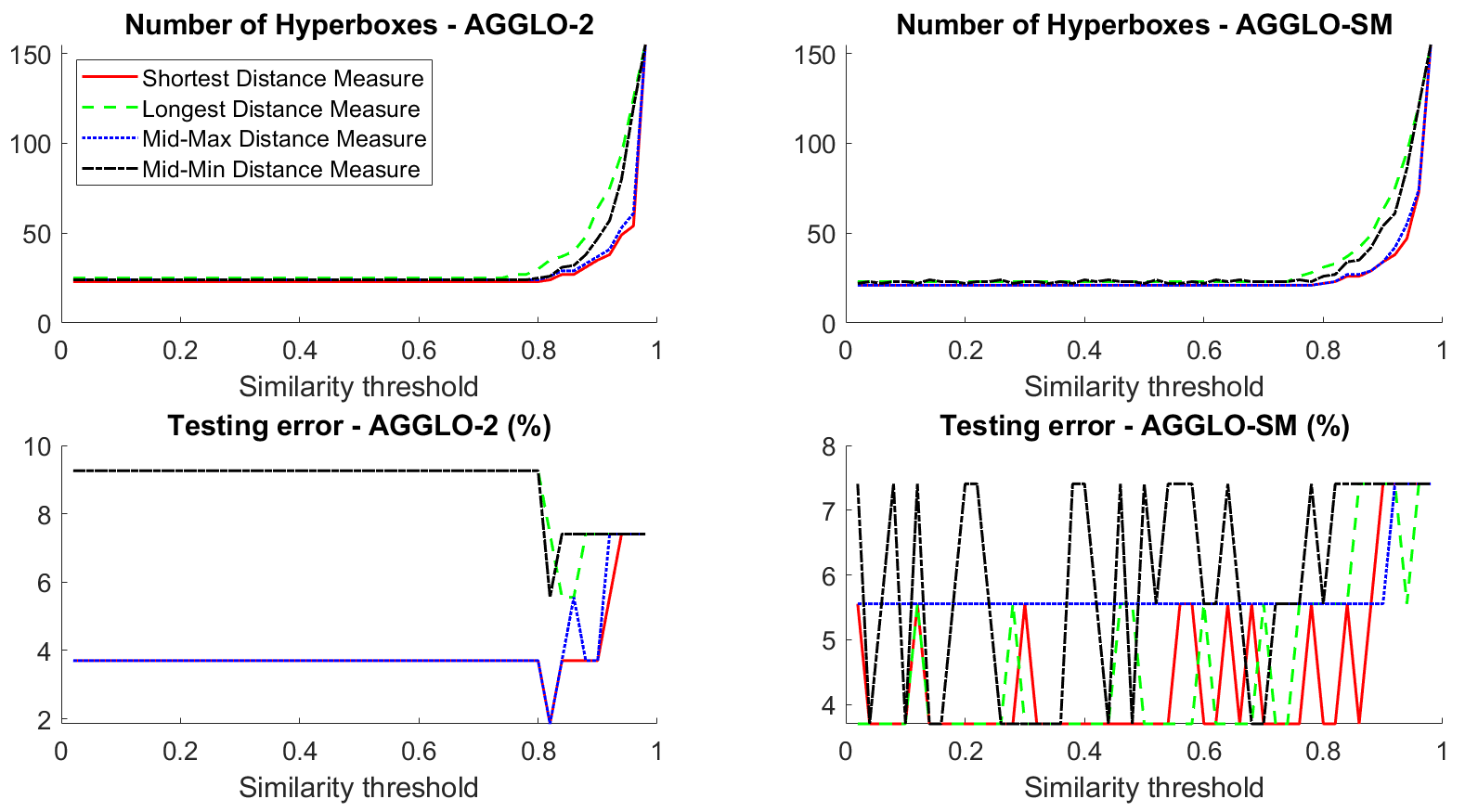}
	\caption{The influence of similarity threshold on the number of hyperboxes and testing errors for GFMM using agglomerative learning on the \textit{Thyroid} dataset}
	\label{fig5}
\end{figure*}

This experiment is to evaluate the influence of the similarity threshold on the performance of AGGLO-2 and AGGLO-SM algorithms using different similarity measures. For each dataset, a fold was selected for testing data, while three other folds were used as training data. The maximum hyperbox size $ \theta = 0.26 $ was used in this experiment. The minimum similarity threshold values ($\sigma$) were moved from 0.02 to 0.98 with the step being 0.02. The graphs showing the change in the number of hyperboxes and the testing error through several typical datasets can be found at \url{https://github.com/UTS-AAi/comparative-gfmm/blob/master/experiment/similarity-threshold-changing.pdf}. An example is presented in Fig. \ref{fig5}.

It can be seen from the figures that the numbers of hyperboxes of both algorithms on all similarity measures regularly increase when the similarity threshold moves to one. Especially, they sharply rise when the threshold is larger than 0.8, and they oscillate a little if the similarity value is less than 0.7. We can see that the number of generated hyperboxes in the case of using the shortest distance measure to compute the similarity degree is lowest, whereas the use of the longest distance measure results in the highest number of generated hyperboxes among four measures.

For the AGGLO-SM algorithm, the selection of the similarity threshold considerably affects the testing error. Its testing error rates oscillate not following a general rule. For the AGGLO-2, the testing error fluctuates only if the value of the similarity threshold is larger than 0.8. Therefore, experiments in the rest of this paper employed a similarity threshold $ \sigma \leq 0.8 $ for the agglomerative learning algorithms. It can be observed that the best performance of the AGGLO-2 algorithm is frequently achieved in the case of using the shortest distance measure. We can recognize that the classification performance of the GFMM using the agglomerative learning algorithms depends on the choice of the similarity measures for each dataset. Of four similarity measures, there is no measure giving the best results on all datasets. Hence, the similarity measure, similarity threshold, and maximum hyperbox size are three hyper-parameters that need to be optimized for each dataset to achieve the best predictive accuracy.

\subsection{Comparison of different versions of GFMM using agglomerative learning}
This part compares the full similarity matrix based agglomerative learning and accelerated agglomerative learning algorithms. Each dataset was split into four folds using the density-preserving sampling method \cite{Budka13}. Each fold was used in turn as testing data, while the remaining folds were employed as the training set. The obtained result of each model is the average result of four testing folds. For a given training set, experiments were repeated ten times to determine the average training time. We established the similarity threshold $ \sigma = 0.8 $ and the maximum hyperbox size $ \theta = 0.26 $ for both algorithms on all datasets. Table \ref{table3} shows the mean values of the number of produced hyperboxes, training time, and testing error rate of each algorithm through typical datasets.

\begin{table*}
\centering
\caption{The comparison of the full similarity matrix based agglomerative learning and accelerated agglomerative learning, $ \theta = 0.26, \sigma = 0.8 $} \label{table3}
\begin{tabular}{|l|l|r|r|r|r|r|r|}
\hline
\multirow{2}{*}{ID} & \multirow{2}{*}{Dataset} & \multicolumn{3}{c|}{AGGLO-2} & \multicolumn{3}{c|}{AGGLO-SM} \\ \cline{3-8} 
 & & No. Hyperboxes & Training time (s) & Testing Error (\%) & No. Hyperboxes & Training time (s) & Testing error (\%) \\ \hline
1  & Circle                 & 40.750  & 0.196  & 3.200       & 41     & 21.998  & 3.300      \\ \hline
2  & Complex 9               & 31.750  & 0.932 & 0.165   & 30.500   & 229.952 & 0.231 \\ \hline
3  & Diagnostic Breast Cancer & 133.500  & 0.579  & 5.6252   & 133.250 & 20.515  & 5.622  \\ \hline
4  & Glass                  & 47.500   & 0.043  & 35.500 & 47.750  & 0.564  & 41.125 \\ \hline
5  & ionosphere             & 151.750 & 0.179  & 11.406 & 152.250 & 3.164  & 11.974 \\ \hline
6  & Iris                   & 18.250  & 0.023 & 4.623  & 17.500   & 0.173  & 4.623  \\ \hline
7  & Segmentation           & 243.750 & 2.237 & 4.285  & 240.750 & 171.512 & 3.982  \\ \hline
8  & Spherical\_5\_2        & 13.750  & 0.029  & 1.197  & 12.750  & 0.639  & 0.397 \\ \hline
9  & Spiral                 & 28.500   & 0.169 & 0.100      & 24.500   & 12.132 & 0        \\ \hline
10 & Thyroid                & 26     & 0.037  & 5.573   & 24.500   & 0.599    & 4.167 \\ \hline
11 & Wine                   & 89.250  & 0.061  & 5.076   & 91     & 0.391   & 5.076  \\ \hline
12 & Yeast                  & 144.250 & 1.295 & 68.661 & 139.750 & 97.463 & 70.348 \\ \hline
13 & Zelnik6                & 12.750  & 0.031  & 0.424  & 12.500   & 0.681  & 0        \\ \hline
\end{tabular}
\end{table*}

As indicated in the table, the AGGLO-2 algorithm is from one to two orders of magnitude faster than the AGGLO-SM in almost all datasets. However, the average number of hyperboxes generated in the AGGLO-2 is slightly higher than that of hyperboxes created by the AGGLO-SM algorithm. The average testing error values of the GFMM neural network using the AGGLO-2 are slightly higher than those using the AGGLO-SM algorithm on many datasets except \textit{Circle}, \textit{Complex 9}, \textit{Glass}, \textit{Ionosphere}, and \textit{Yeast}. In general, the predictive results using the GFMM trained by the AGGLO-2 is relatively the same as those implementing the AGGLO-SM while the training time is much faster. As a result, the AGGLO-2 algorithm significantly improves the performance of the full similarity matrix based agglomerative learning algorithm. It is noted that the training time of the AGGLO-SM algorithm for large-sized training datasets such as \textit{ringnorm}, \textit{twonorm}, and \textit{waveform} is extremely long (more than two days for each iteration), so they were not reported in this paper. The computational expense of the AGGLO-SM is costly because its time complexity is $ O(n^3) $. This fact prevents the applicability of the AGGLO-SM in tackling large-sized datasets. In the rest of this paper, the AGGLO-2 was implemented for the next experiments to compare to other classification algorithms.

\subsection{The influence of data presentation order on the performance of GFMM classifiers}
This experiment is to assess the impact of data presentation order to the classification performance of incremental learning and agglomerative learning algorithms of the GFMM network. For each dataset, one fold was chosen as the testing set, and three remaining folds were training data. Each experiment was executed ten times, and each time randomly shuffled the order of samples in the same training set, and three learning algorithms were trained on the same dataset. We set the similarity threshold $ \sigma = 0.8 $ using the shortest distance measure for the similarity computation and the maximum hyperbox size parameter $ \theta = 0.26 $. Table \ref{table4} reports the standard deviation (std) of the number of hyperboxes and testing errors of different algorithms applied to 13 datasets.

\begin{table*}
\centering
\caption{Standard deviation on results of diffirent versions of GFMMs due to the impact of presentation order} \label{table4}
\begin{tabular}{|l|l|c|c|c|c|c|c|}
\hline
\multirow{2}{*}{ID} & \multirow{2}{*}{Dataset} & \multicolumn{2}{c|}{Online} & \multicolumn{2}{c|}{AGGLO-2} & \multicolumn{2}{c|}{AGGLO-SM} \\ \cline{3-8} 
 & & No. hyperboxes & Testing Error (\%) & No. hyperboxes & Testing Error (\%) & No hyperboxes & Testing Error (\%) \\ \hline
1  & Circle                 & 1.687 & 0.844 & 1.059 & 0.627 & 0           & 0           \\ \hline
2  & Complex 9              & 1.287 & 0.377 & 0.994 & 0.056 & 0 & 0 \\ \hline
3  & Diagnostic Breast Cancer & 3.011 & 1.188 & 2.944 & 0.996 & 1.054 & 0.371 \\ \hline
4  & Glass                  & 1.059 & 5.031 & 0.667 & 3.334 & 0           & 0           \\ \hline
5  & ionosphere             & 1.430 & 1.290 & 1.265 & 0.804 & 0           & 0           \\ \hline
6  & Iris                   & 0.632 & 1.140 & 1.370 & 1.396 & 0.949 & 0           \\ \hline
7  & Segmentation           & 5.446 & 0.364 & 3.736 & 0.390 & 0.516 & 0           \\ \hline
8  & Spherical\_5\_2        & 1.174 & 0.502 & 0.707 & 0.837 & 0           & 0           \\ \hline
9  & Spiral                 & 1.764 & 2.892 & 1.491 & 0           & 0           & 0           \\ \hline
10 & Thyroid                & 1.197    & 1.991 & 0.816 & 1.295 & 0           & 0.895 \\ \hline
11 & Wine                   & 1.829  & 1.174 & 1.633 & 0           & 0           & 0           \\ \hline
12 & Yeast                  & 2.058	& 2.107	& 2.406 & 1.222 & 1.337 & 0.475 \\ \hline
13 & Zelnik6                & 0.667 & 0           & 0.422 & 0           & 0           & 0           \\ \hline
\end{tabular}
\end{table*}

It is seen that the standard deviation values of the testing errors of the GFMM trained by the AGGLO-SM algorithm are zero on almost all datasets, except \textit{Yeast}, \textit{Diagnostic Breast Cancer}, and \textit{Thyroid}. Even on these three datasets, the standard deviation of testing error values is very small ($ < 0.9 \% $). These figures indicate that the full similarity matrix based learning algorithm is almost unaffected by the input data presentation order. In contrast, of three learning algorithms, the incremental learning version is most affected by the data presentation order since hyperboxes are adjusted for each input pattern. The AGGLO-2 is also influenced by the data presentation order because it selects, in turn, each hyperbox to calculate the similarity value with the other ones, but the standard deviation values of testing errors are quite tiny. This experiment confirms that agglomerative learning algorithms are stable against the change of presentation order within training data.

\subsection{Comparison of GFMM and other types of fuzzy min-max neural networks} \label{comfmnn}
This experiment aims to compare the performance of the GFMM networks to other types of fuzzy min-max neural networks using the expansion and contraction phases in the learning algorithm such as the original fuzzy min-max neural network \cite{Simpson92}, the enhanced fuzzy min-max neural network \cite{Mohammed15}, and the enhanced fuzzy min-max neural network with the K-nearest hyperbox selection rule \cite{Mohammed17}.

Through experimental results mentioned above, we have observed that the performance of fuzzy min-max neural networks depends on the value of maximum hyperbox size for each dataset. Therefore, we used the grid search method and 3-fold cross-validation for tuning the maximum hyperbox size of the classification model on validation sets among values within the list of $\theta \in \{0.06, 0.1, 0.16, 0.2, 0.26,\ldots,0.8\}$. In addition to the maximum hyperbox size, the KNEFMNN model also depends on the number of selected hyperboxes ($K$) for the hyperbox expansion process. We set the searching range of K in the range of $[2, 10]$. As for the AGGLO-2 version, we used the longest-distance measure and set the similarity threshold $\sigma = 0$ so that the GFMM model using this agglomerative learning algorithm is only dependent on the value of $\theta$. It is not feasible to exhaustively explore all the possible values for the maximum hyperbox size value, and the purpose of this paper is to compare the performance of the fuzzy min-max classifiers, not on the fine-tuning approaches, so we limited the number of values for each parameter.

Each dataset was split into four folds using the density-preserving sampling method \cite{Budka13}. Each fold was selected as testing set in turn, while three remaining folds were employed as the training and validation data. Assuming that $F_1, F_2$, and $F_3$ are three folds used for parameter-tuned process, we employed $F_1$ and $F_2$ as training data to construct the fuzzy min-max classifiers for each value of $\theta$. Then, the error rate on the validation fold $F_3$ is computed. This process is repeated for $F_1$ and $F_2$ used as the validation set. The value of $\theta$ leading to the lowest averaged prediction error on three folds is selected to build the final fuzzy min-max classifier on the training set containing all $F_1, F_2$ and $F_3$ folds.

Table \ref{table10} shows the mean values of the number of generated hyperboxes, training time, parameter-tuned time, and testing error for each learning algorithm on four testing folds using different datasets. Table \ref{table11} reports the ranks of algorithms in terms of training time, parameter-tuned time, and testing errors.

Regarding training time, it is seen that Simpson's learning algorithm in the FMNN is fastest, while the AGGLO-2 is slowest. The online version of the GFMM executes more rapidly compared to improved versions of the FMNN such as the EFMNN or KNEFMNN. It can be seen that the EFMNN using the K-nearest hyperbox selection runs faster than the EFMNN in some cases, but in general it is slower than the EFMNN with optimized parameters. In terms of parameter-tuned time, the KNEFMNN is slowest in most cases, but on medium-sized datasets such as \textit{Ringnorm}, \textit{Twonorm}, and \textit{Waveform}, the time to find the best parameters of AGGLO-2 is longest. Therefore, the current version of AGGLO-2 algorithm should not be used for tuning parameters in an automatic manner in cases of large-sized training datasets.

The number of hyperboxes generated by the learning algorithms of the GFMM is fewest in general, while the EFMNNN and the original FMNN produce the largest number of hyperboxes. The use of K-nearest hyperbox selection rule in the KNEFMNN also helps considerably reduce the number of hyperboxes created by the EFMNN. We can observe that the GFMM and KNEFMNN generate quite fewer hyperboxes compared to the FMNN or EFMNN since they consider many current hyperboxes for the expansion conditions before creating new hyperboxes. $ K $ hyperboxes are taken into account in the KNEFMNN, and as many hyperboxes as possible are considered in the GFMM network, whereas the FMNN and EFMNN produce a new hyperbox when the winner hyperbox does not meet the expansion constraints.

Generally, the KNEFMNN reduces the number of generated hyperboxes and increases the accuracy of the EFMNN on the considered datasets. The best classification performance belongs to the KNEFMNN, and the online version of GFMM and the EFMNN achieve the worst classification results. We can observe that, on average, only AGGLO-2 and KNEFMNN refine the accuracy of the original FMNN using optimal parameter configurations, but their training time increases substantially. Althout the AGGLO-2 is a promising learning algorithm, its running time is still long on the large datasets. Therefore, many research efforts should be put on improving this algorithm.

It can be easily observed that the number of generated hyperboxes in fuzzy min-max classifiers is large because the best performance of models is achieved for a small value of $\theta$. As shown in the example in Section \ref{secprob}, small values of the maximum hyperbox size result in complex models, which are more likely to overfit the training data. Therefore, to assess the efficiency of hyper-parameters selected using density-preserving and cross-validation methods, we trained the models using the same best parameters returned by grid-search procedure on only two DPS folds instead of three DPS folds as in the above experiments. The remaining fold was used as a validation set to conduct the hyperbox pruning. The hyperboxes with the predictive accuracy on the validation set less than a user-defined threshold (0.5 in this work) were removed. It is noted that there are several hyperboxes that do not take part in the pruning process as they have not been used to classify any validation samples (i.e., they have not been the ``winners"). Therefore, there is no information about their potential predictive accuracy, and they can be pruned or retained. The decision of removing or keeping such hyperboxes depends on the misclassification error of the final model on the validation set. If the removal of these hyperboxes leads to the lower error rates on the validation set, they will be pruned, and vice versa.

\pagebreak
\begin{table*}[!ht]
\centering
\caption{Average performance of different variants of fuzzy min-max neural networks} \label{table10}
\begin{tabular}{|c|l|l|r|r|r|r|r|}
\hline
\textbf{ID}                  & \textbf{Dataset}                                   & \textbf{Measure}              & \textbf{Online GFMM} & \textbf{AGGLO-2}    & \textbf{FMNN}     & \textbf{EFMNN}    & \textbf{KNEFMNN}   \\ \hline
\multirow{4}{*}{1}  & \multirow{4}{*}{Circle}                   & No. of hyperboxes    & 172         & 126.25     & 209.75   & 282.75   & 116.5     \\ \cline{3-8} 
                    &                                           & Training time (s)        & 1.2473      & 3.5819     & 1.682    & 3.6704   & 1.6951    \\ \cline{3-8} 
                    &                                           & Testing error (\%)   & 3.4         & 3.6        & 4.3      & 3.1      & 3.7       \\ \cline{3-8} 
                    &                                           & Parameter-tuned time (s) & 9.913       & 29.0167    & 18.08    & 23.6851  & 155.4245  \\ \hline
\multirow{4}{*}{2}  & \multirow{4}{*}{Complex 9}                & No. of hyperboxes    & 198.75      & 213        & 450.25   & 458.5    & 257.25    \\ \cline{3-8} 
                    &                                           & Training time (s)       & 4.1982      & 3.613      & 7.2618   & 11.6016  & 7.1803    \\ \cline{3-8} 
                    &                                           & Testing error (\%)   & 0           & 0          & 0.033    & 0        & 0         \\ \cline{3-8} 
                    &                                           & Parameter-tuned time (s) & 36.7573     & 40.6123    & 57.938   & 75.0684  & 424.9958  \\ \hline
\multirow{4}{*}{3}  & \multirow{4}{*}{Diagnostic Breast Cancer} & No. of hyperboxes    & 62.25       & 80.75      & 383      & 381.25   & 257.75    \\ \cline{3-8} 
                    &                                           & Training time (s)       & 0.3179      & 2.6611     & 0.4174   & 0.3406   & 1.2007    \\ \cline{3-8} 
                    &                                           & Testing error (\%)   & 4.7463      & 2.987      & 3.1668   & 4.3955   & 4.0443    \\ \cline{3-8} 
                    &                                           & Parameter-tuned time (s) & 10.0033     & 147.3405   & 6.2507   & 12.1373  & 130.3236  \\ \hline
\multirow{4}{*}{4}  & \multirow{4}{*}{Glass}                    & No. of hyperboxes    & 107.25      & 106.25     & 109      & 110.5    & 101.5     \\ \cline{3-8} 
                    &                                           & Training time (s)      & 0.1327      & 1.06       & 0.1203   & 0.172    & 0.1922    \\ \cline{3-8} 
                    &                                           & Testing error (\%)   & 30.3985     & 30.3895    & 27.1225  & 27.5943  & 25.7338   \\ \cline{3-8} 
                    &                                           & Parameter-tuned time (s) & 2.3779      & 6.7931     & 1.8835   & 3.0415   & 27.5698   \\ \hline
\multirow{4}{*}{5}  & \multirow{4}{*}{Ionosphere}               & No. of hyperboxes    & 191.75      & 113        & 208.5    & 229      & 226       \\ \cline{3-8} 
                    &                                           & Training time (s)      & 0.3292      & 5.3567     & 0.2457   & 0.3203   & 0.3514    \\ \cline{3-8} 
                    &                                           & Testing error (\%)   & 12.2585     & 14.2435    & 10.828   & 8.8328   & 8.8328    \\ \cline{3-8} 
                    &                                           & Parameter-tuned time (s) & 7.5723      & 131.0173   & 5.6189   & 9.4672   & 95.3095   \\ \hline
\multirow{4}{*}{6}  & \multirow{4}{*}{Iris}                     & No. of hyperboxes    & 52.25       & 51.75      & 37.5     & 47.75    & 27.5      \\ \cline{3-8} 
                    &                                           & Training time (s)      & 0.05        & 0.4249     & 0.0205   & 0.071    & 0.0515    \\ \cline{3-8} 
                    &                                           & Testing error (\%)   & 5.299       & 5.299      & 3.983    & 5.3165   & 5.3165    \\ \cline{3-8} 
                    &                                           & Parameter-tuned time (s) & 0.9627      & 2.9324     & 0.8019   & 1.0995   & 10.1352   \\ \hline
\multirow{4}{*}{7}  & \multirow{4}{*}{Ringnorm}                 & No. of hyperboxes    & 507.25      & 1,415.25   & 1,899.75 & 2,263.25 & 1,217.50  \\ \cline{3-8} 
                    &                                           & Training time (s)     & 15.0971     & 1,276.87   & 15.4478  & 25.4722  & 25.0148   \\ \cline{3-8} 
                    &                                           & Testing error (\%)   & 13.0405     & 9.311      & 16.027   & 25.4188  & 18.2705   \\ \cline{3-8} 
                    &                                           & Parameter-tuned time (s) & 621.682     & 117,532.42 & 412.5195 & 555.8019 & 5013.0365 \\ \hline
\multirow{4}{*}{8}  & \multirow{4}{*}{Segmentation}             & No. of hyperboxes    & 803.5       & 809.75     & 906      & 1205.25  & 994.5     \\ \cline{3-8} 
                    &                                           & Training time (s)    & 14.8696     & 192.1328   & 11.7049  & 17.0457  & 19.5409   \\ \cline{3-8} 
                    &                                           & Testing error (\%)   & 4.1558      & 3.9825     & 3.506    & 2.2075   & 2.2508    \\ \cline{3-8} 
                    &                                           & Parameter-tuned time (s) & 130.2684    & 736.4805   & 64.8881  & 261.9439 & 1691.2003 \\ \hline
\multirow{4}{*}{9}  & \multirow{4}{*}{Spherical\_5\_2}          & No. of hyperboxes    & 22          & 23.25      & 21.25    & 24.5     & 14.75     \\ \cline{3-8} 
                    &                                           & Training time (s)     & 0.0593      & 0.1074     & 0.038    & 0.0688   & 0.059     \\ \cline{3-8} 
                    &                                           & Testing error (\%)   & 1.2033      & 0.8        & 1.1905   & 1.197    & 1.197     \\ \cline{3-8} 
                    &                                           & Parameter-tuned time (s) & 1.771       & 2.4714     & 1.6866   & 2.1349   & 18.0612   \\ \hline
\multirow{4}{*}{10} & \multirow{4}{*}{Spiral}                   & No. of hyperboxes    & 121.5       & 115.75     & 102.75   & 137.5    & 121.5     \\ \cline{3-8} 
                    &                                           & Training time (s)     & 0.4895      & 1.851      & 0.4994   & 0.9892   & 0.9478    \\ \cline{3-8} 
                    &                                           & Testing error (\%)   & 0           & 0          & 0        & 0        & 0         \\ \cline{3-8} 
                    &                                           & Parameter-tuned time (s) & 7.7798      & 16.4694    & 8.3901   & 13.0277  & 99.7823   \\ \hline
\multirow{4}{*}{11} & \multirow{4}{*}{Thyroid}                  & No. of hyperboxes    & 68.5        & 48         & 95.25    & 96.5     & 108.5     \\ \cline{3-8} 
                    &                                           & Training time (s)      & 0.0863      & 0.4432     & 0.1249   & 0.1393   & 0.1866    \\ \cline{3-8} 
                    &                                           & Testing error (\%)   & 2.315       & 3.7215     & 3.2408   & 3.7125   & 2.778     \\ \cline{3-8} 
                    &                                           & Parameter-tuned time (s) & 1.4175      & 5.2984     & 1.2786   & 1.885    & 16.3993   \\ \hline
\multirow{4}{*}{12} & \multirow{4}{*}{Twonorm}                  & No. of hyperboxes    & 823.75      & 1,134.75    & 5,448.50   & 5,531.75  & 5,384.25   \\ \cline{3-8} 
                    &                                           & Training time (s)     & 27.9473     & 463.9801   & 13.87    & 7.2354   & 13.8077   \\ \cline{3-8} 
                    &                                           & Testing error (\%)   & 4.527       & 4.3378     & 5.1213   & 5.3108   & 4.1623    \\ \cline{3-8} 
                    &                                           & Parameter-tuned time (s) & 615.0026    & 109,086.28 & 371.7325 & 549.3722 & 4,787.5467 \\ \hline
\multirow{4}{*}{13} & \multirow{4}{*}{Waveform}                 & No. of hyperboxes    & 322.75      & 838        & 3220     & 3749.75  & 2757.25   \\ \cline{3-8} 
                    &                                           & Training time (s)      & 11.1249     & 177.3769   & 5.6624   & 1.8178   & 31.3935   \\ \cline{3-8} 
                    &                                           & Testing error (\%)   & 17.88       & 17.76      & 22.52    & 21.36    & 19.88     \\ \cline{3-8} 
                    &                                           & Parameter-tuned time (s) & 305.3155    & 28,641.43  & 160.9124 & 312.5685 & 2,867.3944 \\ \hline
\multirow{4}{*}{14} & \multirow{4}{*}{Wine}                     & No. of hyperboxes    & 46.25       & 25.75      & 39.25    & 74.5     & 27        \\ \cline{3-8} 
                    &                                           & Training time (s)      & 0.0457      & 0.141      & 0.0368   & 0.0732   & 0.0824    \\ \cline{3-8} 
                    &                                           & Testing error (\%)   & 3.952       & 4.5073     & 2.8155   & 5.6313   & 2.8283    \\ \cline{3-8} 
                    &                                           & Parameter-tuned time (s) & 1.8072      & 9.3405     & 1.2843   & 1.8737   & 19.5112   \\ \hline
\multirow{4}{*}{15} & \multirow{4}{*}{Yeast}                    & No. of hyperboxes    & 738.75      & 537.75     & 859.5    & 913.5    & 663       \\ \cline{3-8} 
                    &                                           & Training time (s)      & 5.4222      & 54.0145    & 4.613    & 5.1744   & 8.0922    \\ \cline{3-8} 
                    &                                           & Testing error (\%)   & 49.3938     & 49.2588    & 49.7978  & 47.17    & 46.2265   \\ \cline{3-8} 
                    &                                           & Parameter-tuned time (s) & 44.0484     & 386.3297   & 33.0291  & 63.2031  & 580.8328  \\ \hline
\multirow{4}{*}{16} & \multirow{4}{*}{Zelnik6}                                    & No. of hyperboxes    & 26          & 40.75      & 59       & 45.25    & 34.5      \\ \cline{3-8} 
                    &                                           & Training time (s)     & 0.0426      & 0.3498     & 0.0933   & 0.0976   & 0.0789    \\ \cline{3-8} 
                    &                                           & Testing error (\%)   & 0.4238      & 0.4238     & 0.4238   & 0.4238   & 0.4238    \\ \cline{3-8} 
                    &                                           & Parameter-tuned time (s) & 1.156       & 3.4367     & 1.1349   & 1.2665   & 10.8995   \\ \hline
\end{tabular}
\end{table*}

\scriptsize{
\onecolumn 
\begin{longtable}[!ht]{|l|l|l|r|r|r|r|r|}
\caption{Average performance of different variants of fuzzy min-max neural networks with a pruning procedure} \label{tableprun}\\
\hline
\textbf{ID}                  & \textbf{Dataset}                                 & \textbf{Measure}                           & \textbf{Online GFMM} & \textbf{AGGLO-2}  & \textbf{FMNN}    & \textbf{EFMNN}    & \textbf{KNEFMNN}     \\ \hline
\endhead
\multirow{5}{*}{1}  & \multirow{5}{*}{Circle}                 & No. of hyperboxes before pruning  & 146.75      & 106.75   & 164     & 226.5    & 99.75       \\ \cline{3-8} 
                    &                                         & No. of hyperboxes after pruning   & 124.75      & 90       & 87.75   & 184.75   & 78.25       \\ \cline{3-8} 
                    &                                         & Training time                     & 0.7504      & 2.3950   & 0.9043  & 1.8106   & 1.0028      \\ \cline{3-8} 
                    &                                         & Testing error before pruning (\%) & 3.3         & 3.2      & 3.9     & 3        & 3.6         \\ \cline{3-8} 
                    &                                         & Testing error after pruning (\%)  & 3.3         & 3.8      & 4.1     & 3.3      & 3.8         \\ \hline
\multirow{5}{*}{2}  & \multirow{5}{*}{Complex 9}              & No. of hyperboxes before pruning  & 183         & 196      & 320.75  & 345.5    & 221.5       \\ \cline{3-8} 
                    &                                         & No. of hyperboxes after pruning   & 156         & 195.25   & 160.5   & 191.75   & 156.75      \\ \cline{3-8} 
                    &                                         & Training time                     & 2.9040      & 2.4693   & 4.3682  & 6.7983   & 4.5382      \\ \cline{3-8} 
                    &                                         & Testing error before pruning (\%) & 0           & 0        & 0.033   & 0        & 0           \\ \cline{3-8} 
                    &                                         & Testing error after pruning (\%)  & 0           & 0        & 0.033   & 0        & 0           \\ \hline
\multirow{5}{*}{3}  & \multirow{5}{*}{DiagnosticBreastCancer} & No. of hyperboxes before pruning  & 48.5        & 57.75    & 254     & 254      & 173.5       \\ \cline{3-8} 
                    &                                         & No. of hyperboxes after pruning   & 24          & 35.75    & 43.5    & 56       & 32.5        \\ \cline{3-8} 
                    &                                         & Training time                     & 0.2235      & 1.3171   & 0.2754  & 0.2481   & 0.5979      \\ \cline{3-8} 
                    &                                         & Testing error before pruning (\%) & 5.273       & 5.8013   & 4.2215  & 4.5735   & 4.3965      \\ \cline{3-8} 
                    &                                         & Testing error after pruning (\%)  & 5.2718      & 5.9760   & 4.223   & 5.4528   & 4.3965      \\ \hline
\multirow{5}{*}{4}  & \multirow{5}{*}{Glass}                  & No. of hyperboxes before pruning  & 78.25       & 77.75    & 72.5    & 79.75    & 73.75       \\ \cline{3-8} 
                    &                                         & No. of hyperboxes after pruning   & 42.5        & 41.75    & 32.75   & 63.25    & 47.5        \\ \cline{3-8} 
                    &                                         & Training time                     & 0.0718      & 0.5802   & 0.0707  & 0.0806   & 0.091       \\ \cline{3-8} 
                    &                                         & Testing error before pruning (\%) & 30.407      & 30.3985  & 27.1318 & 26.66    & 28.066      \\ \cline{3-8} 
                    &                                         & Testing error after pruning (\%)  & 35.045      & 34.5735  & 30.381  & 25.725   & 29.446      \\ \hline
\multirow{5}{*}{5}  & \multirow{5}{*}{ionosphere}             & No. of hyperboxes before pruning  & 131.75      & 78.75    & 141.25  & 159      & 156.25      \\ \cline{3-8} 
                    &                                         & No. of hyperboxes after pruning   & 35          & 26.75    & 33.5    & 73       & 72          \\ \cline{3-8} 
                    &                                         & Training time                     & 0.1991      & 2.2911   & 0.1522  & 0.1646   & 0.1818      \\ \cline{3-8} 
                    &                                         & Testing error before pruning (\%) & 14.5343     & 14.2373  & 11.9645 & 9.401    & 9.117       \\ \cline{3-8} 
                    &                                         & Testing error after pruning (\%)  & 14.8185     & 14.2373  & 14.2405 & 11.3898  & 11.1055     \\ \hline
\multirow{5}{*}{6}  & \multirow{5}{*}{Iris}                   & No. of hyperboxes before pruning  & 39.5        & 38.25    & 23.75   & 37       & 21          \\ \cline{3-8} 
                    &                                         & No. of hyperboxes after pruning   & 21.25       & 22       & 6.5     & 15.25    & 11.75       \\ \cline{3-8} 
                    &                                         & Training time                     & 0.0373      & 0.2035   & 0.0219  & 0.0449   & 0.0361      \\ \cline{3-8} 
                    &                                         & Testing error before pruning (\%) & 5.9745      & 4.6585   & 3.983   & 5.9745   & 3.9833      \\ \cline{3-8} 
                    &                                         & Testing error after pruning (\%)  & 5.3165      & 4.641    & 3.983   & 5.9745   & 4.6588      \\ \hline
\multirow{5}{*}{7}  & \multirow{5}{*}{Ringnorm}               & No. of hyperboxes before pruning  & 372.75      & 976.25   & 1132    & 1482.25  & 789.5       \\ \cline{3-8} 
                    &                                         & No. of hyperboxes after pruning   & 207.25      & 716      & 2       & 855.5    & 10          \\ \cline{3-8} 
                    &                                         & Training time                     & 14.0247     & 488.1716 & 9.71927 & 18.79225 & 19.04846739 \\ \cline{3-8} 
                    &                                         & Testing error before pruning (\%) & 12.6758     & 9.9595   & 18.0135 & 26.2163  & 17.4188     \\ \cline{3-8} 
                    &                                         & Testing error after pruning (\%)  & 12.6215     & 9.811    & 18.0135 & 25.7028  & 17.2568     \\ \hline
\multirow{5}{*}{8}  & \multirow{5}{*}{Segmentation}           & No. of hyperboxes before pruning  & 624.75      & 631.75   & 635.25  & 885.25   & 744.5       \\ \cline{3-8} 
                    &                                         & No. of hyperboxes after pruning   & 530.75      & 545.5    & 190.5   & 506.25   & 447.5       \\ \cline{3-8} 
                    &                                         & Training time                     & 7.0460      & 84.3958  & 5.6199  & 7.3270   & 8.4242      \\ \cline{3-8} 
                    &                                         & Testing error before pruning (\%) & 4.8918      & 4.935    & 3.723   & 2.857    & 3.073       \\ \cline{3-8} 
                    &                                         & Testing error after pruning (\%)  & 5.7575      & 5.6278   & 4.632   & 3.7663   & 3.8528      \\ \hline
\multirow{5}{*}{9}  & \multirow{5}{*}{Spherical\_5\_2}        & No. of hyperboxes before pruning  & 19.75       & 19.5     & 15.25   & 18.75    & 13.25       \\ \cline{3-8} 
                    &                                         & No. of hyperboxes after pruning   & 17.25       & 15.25    & 9.5     & 11.25    & 10.75       \\ \cline{3-8} 
                    &                                         & Training time                     & 0.0567      & 0.0934   & 0.0358  & 0.0546   & 0.0485      \\ \cline{3-8} 
                    &                                         & Testing error before pruning (\%) & 1.197       & 1.6003   & 1.5875  & 2.0035   & 2.4068      \\ \cline{3-8} 
                    &                                         & Testing error after pruning (\%)  & 1.197       & 1.197    & 1.58725 & 2.4003   & 2.4068      \\ \hline
\multirow{5}{*}{10} & \multirow{5}{*}{Spiral}                 & No. of hyperboxes before pruning  & 103         & 105      & 81.5    & 109.25   & 103         \\ \cline{3-8} 
                    &                                         & No. of hyperboxes after pruning   & 92          & 105      & 69.75   & 95.25    & 94.5        \\ \cline{3-8} 
                    &                                         & Training time                     & 0.3895      & 1.3104   & 0.3818  & 0.6423   & 0.6513      \\ \cline{3-8} 
                    &                                         & Testing error before pruning (\%) & 0           & 0        & 0       & 0        & 0           \\ \cline{3-8} 
                    &                                         & Testing error after pruning (\%)  & 0           & 0        & 0       & 0        & 0           \\ \hline
\multirow{5}{*}{11} & \multirow{5}{*}{Thyroid}                & No. of hyperboxes before pruning  & 53          & 35       & 65.25   & 68.75    & 77.75       \\ \cline{3-8} 
                    &                                         & No. of hyperboxes after pruning   & 36          & 21.25    & 18.75   & 24.75    & 31.5        \\ \cline{3-8} 
                    &                                         & Training time                     & 0.0547      & 0.2263   & 0.0694  & 0.0627   & 0.0838      \\ \cline{3-8} 
                    &                                         & Testing error before pruning (\%) & 3.241       & 4.6475   & 5.5643  & 2.7868   & 2.315       \\ \cline{3-8} 
                    &                                         & Testing error after pruning (\%)  & 3.2408      & 6.036    & 6.0273  & 4.6475   & 3.7128      \\ \hline
\multirow{5}{*}{12} & \multirow{5}{*}{Twonorm}                & No. of hyperboxes before pruning  & 609.75      & 776.25   & 3655    & 3694.5   & 3563.75     \\ \cline{3-8} 
                    &                                         & No. of hyperboxes after pruning   & 315.25      & 610.25   & 2864.25 & 3048     & 27          \\ \cline{3-8} 
                    &                                         & Training time                     & 23.5492     & 215.1739 & 13.3195 & 10.2797  & 15.0206     \\ \cline{3-8} 
                    &                                         & Testing error before pruning (\%) & 4.7703      & 4.108    & 5.297   & 5.3648   & 4.4865      \\ \cline{3-8} 
                    &                                         & Testing error after pruning (\%)  & 4.8378      & 4.2973   & 5.4728  & 5.189    & 4.4865      \\ \hline
\multirow{5}{*}{13} & \multirow{5}{*}{Waveform}               & No. of hyperboxes before pruning  & 247.25      & 565.75   & 2153.75 & 2500     & 1751.5      \\ \cline{3-8} 
                    &                                         & No. of hyperboxes after pruning   & 208.25      & 402.25   & 603     & 2354.75  & 46.25       \\ \cline{3-8} 
                    &                                         & Training time                     & 10.0276     & 85.5954  & 6.3840  & 4.4676   & 23.4186     \\ \cline{3-8} 
                    &                                         & Testing error before pruning (\%) & 19.48       & 18.84    & 22.82   & 20.48    & 20          \\ \cline{3-8} 
                    &                                         & Testing error after pruning (\%)  & 19.36       & 18.4     & 22.6    & 19.7     & 19.66       \\ \hline
\multirow{5}{*}{14} & \multirow{5}{*}{Wine}                   & No. of hyperboxes before pruning  & 31.5        & 20.5     & 28      & 51       & 20          \\ \cline{3-8} 
                    &                                         & No. of hyperboxes after pruning   & 28          & 13.75    & 5.25    & 8.25     & 6.75        \\ \cline{3-8} 
                    &                                         & Training time                     & 0.0373      & 0.1165   & 0.0312  & 0.0480   & 0.0483      \\ \cline{3-8} 
                    &                                         & Testing error before pruning (\%) & 3.9268      & 3.9268   & 2.8155  & 5.0883   & 3.9268      \\ \cline{3-8} 
                    &                                         & Testing error after pruning (\%)  & 3.9268      & 3.9268   & 2.8155  & 5.0883   & 3.9268      \\ \hline
\multirow{5}{*}{15} & \multirow{5}{*}{Yeast}                  & No. of hyperboxes before pruning  & 522         & 387      & 582.5   & 618      & 461.25      \\ \cline{3-8} 
                    &                                         & No. of hyperboxes after pruning   & 267.25      & 220.25   & 416.5   & 443.25   & 350.25      \\ \cline{3-8} 
                    &                                         & Training time                     & 2.3044      & 21.8444  & 1.9543  & 2.1734   & 3.3138      \\ \cline{3-8} 
                    &                                         & Testing error before pruning (\%) & 49.7305     & 49.8655  & 51.1455 & 47.6415  & 46.5633     \\ \cline{3-8} 
                    &                                         & Testing error after pruning (\%)  & 49.5283     & 47.5068  & 47.9783 & 44.6765  & 45.3505     \\ \hline
\multirow{5}{*}{16} & \multirow{5}{*}{Zelnik6}                & No. of hyperboxes before pruning  & 23          & 35.25    & 42      & 36.25    & 29.75       \\ \cline{3-8} 
                    &                                         & No. of hyperboxes after pruning   & 16.5        & 24.25    & 23      & 25.25    & 20          \\ \cline{3-8} 
                    &                                         & Training time                     & 0.0417      & 0.2616   & 0.0565  & 0.0557   & 0.0539      \\ \cline{3-8} 
                    &                                         & Testing error before pruning (\%) & 0.8475      & 1.2643   & 0.8475  & 0.8475   & 0.8475      \\ \cline{3-8} 
                    &                                         & Testing error after pruning (\%)  & 1.695       & 3.3618   & 2.1045  & 2.0975   & 1.688       \\ \hline
\end{longtable}
}

\twocolumn
\normalsize
Table \ref{tableprun} shows results before and after applying the pruning procedure. The model trained on two DPS folds was verified on the same testing sets as in the previous experiment. It can be seen that the number of hyperboxes after performing the pruning operation is significantly reduced. The pruning procedure contributes to small reduction of the classification errors on four datasets, keeping the same errors on four datasets, and slightly increasing error rates ($< 2\%$) on eight datasets. These outcomes show that the learning algorithms using best hyper-parameters and training sets generated by the density-preserving sampling method produced the nearly optimal decision boundaries. In such cases, it has been observed that the pruning process can have a small negative effect and can lead to the increase of the testing errors. However, the validation set is also representative of the underlying data distribution, so the error only grows a little. Only for the \textit{Glass} dataset, the error rate increases by around 5\% after conducting the pruning operation. This case can be explained by the unrepresentative of the validation set. This dataset has a small number of patterns, while it has a high number of features and classes. Therefore, the samples are sparsely distributed in the input space, and the DPS method may not find the representative subsets. In general, the error rates of models trained on two DPS folds are slightly higher than those of classifiers trained on three DPS folds. These results confirm that the DPS method generated representative subsets for small datasets to assist the learning algorithms. The obtained results also indicate that the overfitting phenomenon on the training set does not always result in the bad predictive performance on unseen data if the training data are representative patterns of the underlying data distribution.

To better understand the performance of fuzzy min-max neural networks, a rigorous statistical significance test procedure will be employed to interpret the obtained results on the considered datasets. We only perform statistical testing for results of classifiers trained on whole training sets. Our null hypothesis is:

$ H_0 $: \textit{There is no difference in the performance of different types of fuzzy min-max neural networks on 16 different experimental datasets} 

To reject this hypothesis, we will use a ``multiple testing" procedure. Two methods regularly used to test the significant differences among multiple samples are a parametric analysis of variance (ANOVA) and its non-parametric counterparts such as the Friedman test. In a survey on the theoretical work of statistical tests, Demsar \cite{Demsar06} recommended that the Friedman test with a relevant posthoc test should be utilized in the case of the comparisons conducted on more than two objects. This paper employs the Friedman rank-sum test \cite{Eisinga17} to evaluate the classification performance statistically because the testing error values of predictors do not follow any symmetric distribution. Firstly, the Friedman rank-sum test ranks the performance of classification algorithms with the best classifier assigned the first rank, and the second best ranked two, etc. Then, the Friedman test performs comparisons on the average ranks of classifiers. Table \ref{table11} shows ranks over five learning algorithms of different types of fuzzy min-max neural networks as well as the average rank on 16 datasets.

Let $ r_i^{j} $ be the rank of the $ j^{th} $ model in $ k $ models on the $ i^{th} $ dataset of $ N $ datasets, where $ k $ is equal to 5 and $ N $ is 16 in this experiment. A null hypothesis as mentioned above states that all algorithms perform similarly, so their average ranks $ R_j $ should be equal, and the Friedman statistic

\begin{equation}
\chi_F^2 = \frac{12 \cdot N}{k \cdot (k + 1)} \left[\sum_{j}^{} {R_j^2 - \frac{k \cdot (k + 1)^2}{4}} \right]
\end{equation}
is distributed according to $ \chi_F^2 $ with $ k - 1 $ degrees of freedom when $ N $ and $ k $ are big enough, i.e., $ N \geq 10  $ and $ k \geq 5 $. Nonetheless, Iman and Davenport \cite{Iman80} claimed that Friedman's $ \chi_F^2 $ is undesirably conservative, and they introduced a better new statistic:

\begin{equation}
F_F = \frac{(N- 1) \cdot \chi_F^2}{N \cdot (k - 1) - \chi_F^2}
\end{equation}
This metric is distributed according to the F-distribution with $ k - 1 $ and $ (k - 1) \cdot (N - 1) $ degrees of freedom. If the null hypothesis is rejected, i.e., the performances of fuzzy min-max neural networks are statistically different, a posthoc test needs to be carried out to find the critical difference among the average ranks of those models.

\begin{table*} [!ht]
\caption{Ranking of the different FMNN variants} \label{table11}
\scriptsize{
\begin{tabular}{|c|p{1.5cm}|p{0.65cm}|p{0.6cm}|p{0.6cm}|p{0.65cm}|p{0.6cm}|p{0.6cm}|p{0.65cm}|p{0.6cm}|p{0.6cm}|p{0.65cm}|p{0.6cm}|p{0.6cm}|p{0.65cm}|p{0.6cm}|p{0.6cm}|}
\hline
\multirow{2}{*}{ID} & \multirow{2}{*}{Dataset} & \multicolumn{3}{c|}{Online GFMM} & \multicolumn{3}{c|}{AGGLO-2}  & \multicolumn{3}{c|}{FMNN}     & \multicolumn{3}{c|}{EFMNN}    & \multicolumn{3}{c|}{KNEFMNN} \\ \cline{3-17} 
 &  & Training time  & Para-tuned time  & Testing error  & Training time & Para-tuned time & Testing error & Training time & Para-tuned time & Testing error & Training time & Para-tuned time& Testing error & Training time  & Para-tuned time & Testing error     \\ \hline
1     & Circle                     & 1      & 1      & 2    & 4      & 4      & 3      & 2      & 2    & 5     & 5      & 3      & 1    & 3     & 5      & 4      \\ \hline
2     & Complex9                   & 2      & 1      & 2.5  & 1      & 2      & 2.5    & 4      & 3    & 5     & 5      & 4      & 2.5  & 3     & 5      & 2.5    \\ \hline
3     & Diagnostic Breast Cancer   & 1      & 2      & 5    & 5      & 5      & 1      & 3      & 1    & 2     & 2      & 3      & 4    & 4     & 4      & 3      \\ \hline
4     & Glass                      & 2      & 2      & 5    & 5      & 4      & 4      & 1      & 1    & 2     & 3      & 3      & 3    & 4     & 5      & 1      \\ \hline
5     & Ionsphere                  & 3      & 2      & 4    & 5      & 5      & 5      & 1      & 1    & 3     & 2      & 3      & 1.5  & 4     & 4      & 1.5    \\ \hline
6     & Iris                       & 2      & 2      & 2.5  & 5      & 4      & 2.5    & 1      & 1    & 1     & 4      & 3      & 4.5  & 3     & 5      & 4.5    \\ \hline
7     & Ringnorm                   & 2      & 3      & 2    & 5      & 5      & 1      & 3      & 1    & 3     & 5      & 2      & 5    & 4     & 4      & 4      \\ \hline
8     & Segmentation               & 2      & 2      & 5    & 5      & 4      & 4      & 1      & 1    & 3     & 3      & 3      & 1    & 4     & 5      & 2      \\ \hline
9     & Spherical\_5\_2            & 3      & 2      & 5    & 5      & 4      & 1      & 1      & 1    & 2     & 4      & 3      & 3.5  & 2     & 5      & 3.5    \\ \hline
10    & Spiral                     & 1      & 1      & 3    & 5      & 4      & 3      & 2      & 2    & 3     & 4      & 3      & 3    & 3     & 5      & 3      \\ \hline
11    & Thyroid                    & 1      & 2      & 1    & 5      & 4      & 5      & 2      & 1    & 3     & 3      & 3      & 4    & 4     & 5      & 2      \\ \hline
12    & Twonorm                    & 4      & 3      & 3    & 5      & 5      & 2      & 3      & 1    & 4     & 1      & 2      & 5    & 2     & 4      & 1      \\ \hline
13    & Waveform                   & 3      & 2      & 2    & 5      & 5      & 1      & 2      & 1    & 5     & 1      & 3      & 4    & 4     & 4      & 3      \\ \hline
14    & Wine                       & 2      & 2      & 3    & 5      & 4      & 4      & 1      & 1    & 1     & 3      & 3      & 5    & 4     & 5      & 2      \\ \hline
15    & Yeast                      & 3      & 2      & 4    & 5      & 4      & 3      & 1      & 1    & 5     & 2      & 3      & 2    & 4     & 5      & 1      \\ \hline
16    & Zelnik6                    & 1      & 2      & 3    & 5      & 4      & 3      & 3      & 1    & 3     & 4      & 3      & 3    & 2     & 5      & 3      \\ \hline
\multicolumn{2}{|l|}{\textbf{Average rank}} & 2.0625 & 1.9375 & 3.25 & 4.6875 & 4.1875 & 2.8125 & 1.9375 & 1.25 & 3.125 & 3.1875 & 2.9375 & 3.25 & 3.375 & 4.6875 & 2.5625 \\ \hline
\end{tabular}
}
\end{table*}

This paper uses the 95\% confidence interval ($ \alpha = 0.05 $) as a threshold to identify the statistic significance of fuzzy min-max neural networks. Firstly, the Friedman test calculates the F-distribution:

\begin{strip}
\begin{equation*}
\chi_F^2 = \frac{12 \cdot 16}{5 \cdot (5 + 1)} \left[(3.25^2 + 2.8125^2 + 3.125^2 + 3.25^2 + 2.5625^2) - \frac{5 \cdot (5 + 1)^2}{4} \right] = 2.35
\end{equation*} 

\begin{equation*}
F_F = \frac{(16 - 1) \cdot \chi_F^2}{16 \cdot (5 - 1) - \chi_F^2} = \frac{(16 - 1) \cdot 2.35}{16 \cdot (5 - 1) - 2.35} = 0.5718
\end{equation*} 
\end{strip}

With 16 datasets and five classifiers, $ F_F $ is distributed according to the F-distribution with $ 5 - 1 = 4 $ and $ (5 - 1) \cdot (16 - 1) = 60 $ degrees of freedom. The critical value of $ F(4, 60) $ for the significance level $ \alpha = 0.05 $ is 2.5252. It is observed that $ F_F < F(4, 60) $, so the null hypothesis is not rejected. It means that there is no statically significant difference in the performance between the general fuzzy min-max neural network and other types of fuzzy min-max neural networks on the considered datasets.

\subsection{Comparison of GFMM and other machine learning algorithms}
\begin{table*}
\centering
\caption{Comparison of the average testing errors of the GFMM with other machine learning algorithms} \label{table12}
\begin{tabular}{|c|l|r|r|r|r|r|r|}
\hline
\textbf{ID} & \textbf{Dataset}                  & \textbf{Online GFMM} & \textbf{AGGLO-2} & \textbf{KNN}    & \textbf{SVM}    & \textbf{Decision tree} & \textbf{Naive Bayes} \\ \hline
1  & Circle                 & 3.4         & 3.6      & 2.8     & 1.1     & 4.1      & 5.7         \\ \hline
2  & Complex9               & 0           & 0        & 0       & 0       & 0.5613   & 5.279       \\ \hline
3  & DiagnosticBreastCancer & 4.7463      & 2.987    & 2.2848  & 2.11025 & 8.6083   & 6.5018      \\ \hline
4  & Glass                  & 30.3985     & 30.3895  & 28.5028 & 24.7643 & 31.3068  & 52.3933     \\ \hline
5  & Ionsphere              & 12.2585     & 14.2435  & 12.2485 & 4.271   & 10.8088  & 11.1025     \\ \hline
6  & Iris                   & 5.299       & 5.299    & 3.325   & 2.6495  & 5.3343   & 4.641       \\ \hline
7  & Ringnorm               & 13.0405     & 9.311    & 23.2298 & 1.2703  & 11.2298  & 1.3378      \\ \hline
8  & Segmentation           & 4.1558      & 3.9825   & 3.4628  & 2.4675  & 3.3768   & 20.173      \\ \hline
9  & Spherical\_5\_2        & 1.2033      & 0.8      & 2.0033  & 1.6003  & 0.3968   & 1.5875      \\ \hline
10 & Spiral                 & 0           & 0        & 0       & 0       & 0.1      & 34.6        \\ \hline
11 & Thyroid                & 2.315       & 3.7215   & 4.1758  & 3.7128  & 5.1103   & 2.7868      \\ \hline
12 & Twonorm                & 4.527       & 4.33775  & 2.3918  & 2.189   & 15.1215  & 2.108       \\ \hline
13 & Waveform               & 17.88       & 17.76    & 13.9    & 12.74   & 23.24    & 18.96       \\ \hline
14 & Wine                   & 3.952       & 4.50725  & 3.38375 & 1.12375 & 10.07575 & 1.69175     \\ \hline
15 & Yeast                  & 49.3938     & 49.25875 & 40.027  & 37.938  & 43.8005  & 88.342      \\ \hline
16 & Zelnik6                & 0.4238      & 0.4238   & 1.688   & 0       & 0.8405   & 0           \\ \hline
\end{tabular}
\end{table*}

\begin{table*}
\centering
\caption{Ranking of GFMM and other machine learning algorithms} \label{table13}
\begin{tabular}{|c|l|c|c|c|c|c|c|}
\hline
\textbf{ID}     & \textbf{Dataset}                   & \textbf{Online GFMM} &\textbf{ AGGLO-2} & \textbf{KNN}    & \textbf{SVM}     & \textbf{Decision tree}     & \textbf{Naive Bayes} \\ \hline
1      & Circle                    & 3           & 4       & 2      & 1       & 5      & 6           \\ \hline
2      & Complex9                  & 2.5         & 2.5     & 2.5    & 2.5     & 5      & 6           \\ \hline
3      & DiagnosticBreastCancer    & 4           & 3       & 2      & 1       & 6      & 5           \\ \hline
4      & Glass                     & 4           & 3       & 2      & 1       & 5      & 6           \\ \hline
5      & Ionsphere                 & 5           & 6       & 4      & 1       & 2      & 3           \\ \hline
6      & Iris                      & 4.5         & 4.5     & 2      & 1       & 6      & 3           \\ \hline
7      & Ringnorm                  & 5           & 3       & 6      & 1       & 4      & 2           \\ \hline
8      & Segmentation              & 5           & 4       & 3      & 1       & 2      & 6           \\ \hline
9      & Spherical\_5\_2           & 3           & 2       & 6      & 5       & 1      & 4           \\ \hline
10     & Spiral                    & 2.5         & 2.5     & 2.5    & 2.5     & 5      & 6           \\ \hline
11     & Thyroid                   & 1           & 4       & 5      & 3       & 6      & 2           \\ \hline
12     & Twonorm                   & 5           & 4       & 3      & 2       & 6      & 1           \\ \hline
13     & Waveform                  & 4           & 3       & 2      & 1       & 6      & 5           \\ \hline
14     & Wine                      & 4           & 5       & 3      & 1       & 6      & 2           \\ \hline
15     & Yeast                     & 5           & 4       & 2      & 1       & 3      & 6           \\ \hline
16     & Zelnik6                   & 3.5         & 3.5     & 6      & 1.5     & 5      & 1.5         \\ \hline
\multicolumn{2}{|l|}{\textbf{Average rank}} & 3.8125      & 3.625   & 3.3125 & 1.6563 & 4.5625 & 4.0313     \\ \hline
\end{tabular}
\end{table*}

This experiment is to compare the classification performance of the GFMM with other prevalent machine algorithms such as Naive Bayes, K-Nearest neighbors, Support vector machines, and Decision trees. These algorithms were implemented by using the scikit-learn toolbox \cite{Pedregosa11} in Python. Similarly to the above experiments, each dataset was also split into four folds using the density-preserving sample technique. Experiments were conducted on each fold as the testing set in turn and three training and validation folds. The validation fold was used to select the parameters leading to the best performance among a range of setting values for each dataset. This process was mentioned in subsection \ref{comfmnn}. The configuration parameters for GFMM using incremental and AGGLO-2 learning algorithms were remained unchanged as shown in subsection \ref{comfmnn}. As for the value $ K $ of the KNN classifier, we attempted to find the best value in the range of $ [3, 30] $. In terms of decision tree models, we adjusted the tree depth parameter (\textit{max\_depth}) ranging from 3 to 30 and unlimited values. For support vector machines, we used a Radial Basis function (RBF) kernel. There are two parameters needing to adjust for RBF kernel ,i.e., the penalty parameter ($C$) and the parameter gamma ($\gamma$). As shown in \cite{Hsu03}, we set $ C \in \{2^{-5}, 2^{-3}, \ldots, 2^{15}\}$ and $ \gamma \in \{2^{-15}, 2^{-13}, \ldots, 2^3\}$. The Gaussian Naive Bayes model has no hyperparameters, so we used its default settings in the scikit-learn library.

Table \ref{table12} shows the average values of the testing error of different algorithms on four testing folds using the best parameter configurations for each learning model, while Table \ref{table13} reports the ranks among algorithms.

As indicated in Table \ref{table13}, the best algorithm is SVM, followed by KNN. The highest testing error values belong to the decision trees. The AGGLO-2 algorithm outperforms Gaussian Naive Bayes, decision trees, and the incremental learning algorithm, but it cannot overcome the performances of KNN and SVM in general. These results show that the GFMM neural network is competitive to other popular learning models. However, the training and parameter-tuned time of the online and agglomerative learning algorithms of the GFMM classifier is costly compared to other machine learning algorithms. Therefore, the learning algorithms of the GFMM model need to be enhanced in many aspects to deal with the massive datasets.

Although the average performance ranks of the AGGLO-2 and incremental learning algorithms are not the best ones among learning models, we need to assess the level of differences among obtained results in terms of statistical significance. Similarly to statistical hypothesis tests mentioned above, we have a null hypothesis in this experiment:

$ H_0 $: \textit{There is no difference in the performance of the general fuzzy min-max neural network and popular machine learning algorithms on 16 different experimental datasets}

We compute the value of F-distribution as follows:

\begin{strip}
\begin{equation*}
\chi_F^2 = \frac{12 \cdot 16}{6 \cdot (6 + 1)} \left[(3.8125^2 + 3.625^2 + 3.3125^2 + 1.6563^2 + 4.5625^2 + 4.0313^2) - \frac{6 \cdot (6 + 1)^2}{4} \right] = 22.6722
\end{equation*} 
\begin{equation*}
F_F = \frac{(16 - 1) \cdot \chi_F^2}{16 \cdot (6 - 1) - \chi_F^2} = \frac{(16 - 1) \cdot 22.6722}{16 \cdot (6 - 1) - 22.6722} = 5.9323
\end{equation*} 
\end{strip}

With 16 datasets and six classification algorithms, $ F_F $ is distributed according to the F-distribution with $ 6 - 1 = 5 $ and $ (6 - 1) \cdot (16 - 1) = 75 $ degrees of freedom. The critical value of $ F(5, 75) $ for the significance level $ \alpha = 0.05 $ is 2.3366. It is observed that $ F_F > F(5, 75) $, so the null hypothesis is rejected at a high level of significance. Based on these outcomes, we may state that there are statistical differences in the performance of the general fuzzy min-max neural network and popular machine learning algorithms.

A post-hoc test is implemented to verify the significant differences of the incremental and agglomerative learning algorithms and other machine learning models. The post-hoc test used in this study is a step down Holm procedure \cite{Holm79}. The Holm procedure tunes the value of significance level ($ \alpha $) according to a step-down method. Let $ p_1, p_2,...,p_{k - 1} $ be the ordered p-values such that $ p_1 \leq p_2 \leq ... \leq p_{k - 1} $ and $ H_1, H_2,..., H_{k - 1} $ be the respective null hypotheses, the Holm procedure rejects null hypotheses $ H_1 $ to $ H_{i - 1} $ if $ i $ is the smallest integer such that $ p_i > \frac{\alpha}{k - i} $ ($ \alpha = 0.05 $ in this paper). To find the value of $ p_i $ for each pair of predictors, we have to identify the values of $ z_i $ in Eq. \ref{zi}.

\begin{equation}
\label{zi}
z_i = \frac{R_i - R_j}{\sqrt{\frac{k \cdot (k + 1)}{6 \cdot N}}}
\end{equation}
where $ i $ is the control classifier (AGGLO-2 or online GFMM), and $ j $ is the another classifier used in the comparisons, $ R_i $ and $ R_j $ are the average ranks of learners $ i $ and $ j $ respectively. The probability value of $ p_i $ is computed from the corresponding value of $ z_i $ following the normal distribution N(0, 1). The calculating outcomes of the Holm procedure are shown in Table \ref{holm-agglo} for AGGLO-2 and in Table \ref{holm-oln} for incremental learning based GFMM.

\begin{table}[!ht]
	\centering
	\caption{Outcomes of Holm post-hoc test for AGGLO-2}
	\small {
		\begin{tabular}{|l|l|c|c|c|}
			\hline  
			$ i $ & AGGLO-2 vs. & $ z_i $ & $ p_i $ & $ \cfrac{\alpha}{k - i} $\\ 
			\hline 
			1 & SVM & 2.9764 & 0.0029 & 0.01 \\
			2 & Decision tree & -1.4174 & 0.1564 & 0.0125 \\
			3 & Naive Bayes & -0.6143 & 0.5390 & 0.0167 \\
			4 & KNN & 0.4725 & 0.6366 & 0.025 \\
			5 & Online GFMM & -0.2835 & 0.7768 & 0.05 \\
			\hline
		\end{tabular} 
	}
	\label{holm-agglo}
\end{table}

\begin{table}[!ht]
	\centering
	\caption{Outcomes of Holm post-hoc test for incremental learning based GFMM}
	\small {
		\begin{tabular}{|l|l|c|c|c|}
			\hline  
			$ i $ & Online GFMM vs. & $ z_i $ & $ p_i $ & $ \cfrac{\alpha}{k - i} $\\ 
			\hline 
			1 & SVM & 3.2599  & 0.0011 & 0.01 \\
			2 & Decision tree & -1.1339 & 0.2568 & 0.0125 \\
			3 & KNN & 0.7559 & 0.4497 & 0.0167 \\
			4 & Naive Bayes & -0.3308 & 0.7408 & 0.025 \\
			5 & AGGLO-2 & 0.2835 & 0.7768 & 0.05 \\
			\hline
		\end{tabular} 
	}
	\label{holm-oln}
\end{table}

From Tables \ref{holm-agglo} and \ref{holm-oln}, it can be observed that $ i = 2 $ is the smallest integer such that $ p_i > \frac{\alpha}{k - i} $. Therefore, $ H_1 $ is rejected, while null hypotheses $ H_2 $, $ H_3 $, $ H_4 $, and $ H_5 $ are retained. Therefore, AGGLO-2 and incremental learning based GFMM are significantly different from SVM, but there are no statistically significant differences among AGGLO-2, decision tree, Naive Bayes, KNN, and the online version of GFMM at an alpha level of 0.05. These outcomes also indicate that SVM using optimal parameter settings is the best model among considered classifiers. Apart from SVM, learning algorithms of GFMM are competitive to popular machine learning models.

\section{Discussion and research directions}\label{direct}
\subsection{Discussion}
In this part, we highlight several notable issues when conducting a comparative study as follows:

\begin{itemize}
    \item \textbf{The impact of hyper-parameters}: Similarly to other machine learning algorithms, the performance of the hyperbox-based classifiers is also dependent on the selection of hyper-parameters, e.g., maximum hyperbox size, etc. Each training dataset needs specific parameters, and we should not use a fixed setting for all datasets. The selection of suitable hyper-parameters should be conducted by combining k-fold cross-validation and sampling methods. The quality of selected hyper-parameters depends mainly on the quality of the training and validation sets. In general, the DPS method helps to preserve the data density and the classes shapes, so the performance of the model trained on small number of DPS folds is not significantly different in comparison to one trained on all DPS folds.
    
    \item \textbf{Selection of training and validation sets}: Experimental results confirm the crucial roles of the choice of training and validation data. If we can build a training set which is representative of the overall data distribution for a given problem, a model which overfits on the training sets still performs well on the testing set. The use of the density-preserving sampling method contributes to forming such representative training samples with nearly the same distribution as the whole dataset. The average testing error rates through different density-preserving sampling folds can be used as the generalization error of the model. Therefore, the hyper-parameters which lead to the lowest error rates on different DPS validation folds may form a trained hyperbox-based classifier with nearly optimal decision boundaries. It is also noted that a model trained on many representative patterns usually achieves higher accuracy than the model trained on a lower number of representative samples. However, if the training sets do not reflect the data density distribution accurately or the constructed model is too complicated, one needs to use overfitting prevention methods.
    
    \item \textbf{Overfitting prevention mechanisms}: Training model with more relevant and clean data is one of the approaches to restrict the negative impact of overfitting. In practice, however, it is difficult to gather many clean training samples. For a small number of training patterns such as datasets in this paper, cross-validation and density-preserving sampling, which are the most appropriate methods, allow us to select the best set of hyper-parameters. In some cases, the best hyper-parameters can lead to complex models and make generalization error increase because of its overfitting on the training set. Therefore, several overfitting prevention techniques such as pruning should be used to eliminate low-quality hyperboxes. However, this method does not always work for all cases. If the training set is representative of underlying data distribution and the best-selected hyper-parameters form a nearly optimal decision boundary, the pruning operation is more likely to cause the loss of some critical information and increase testing error. In addition, the efficiency of the pruning procedure mainly depends on the quality of validation sets. In the case of sparse data with high dimensionality, a high number of classes, and a low number of samples, the DPS method cannot return the representative datasets, so the pruning operation can result in considerable increase of the testing error rates.
\end{itemize}

\subsection{Research directions}
Through experimental results, it can be easily observed that the performance of the incremental learning version of the GFMM neural network depends considerably on the value of maximum hyperbox size threshold. For the agglomerative learning algorithms, apart from the maximum hyperbox size threshold, they also depend on the similarity measures and the minimum similarity threshold. Another parameter also makes an impact on the performance of learning algorithms, but it is not yet considered in this paper. It is parameter $ \gamma $ in the membership function. To find the best values of hyper-parameters for each algorithm, therefore, automatic methods need to be deployed.

In terms of statistical significance, the agglomerative learning algorithm has not shown the significant difference in the predictive accuracy over considered datasets in comparison to the improved incremental learning variants of the fuzzy min-max neural network. In contrast, the training time of the agglomerative learning algorithms, especially the full similar matrix-based algorithm (AGGLO-SM), is much slower than the incremental learning algorithms. The high computational expense will interfere with the applicability of the agglomerative learning algorithms to pattern recognition problems using big data. Furthermore, the performance of the agglomerative learning versions has not outperformed the popular machine learning algorithms, especially SVM. Hence, we need to enhance the efficiency of agglomerative learning algorithms in terms of running time and accuracy or using them for appropriate parts of the learning process. One of the directions to accelerate the training time is the use of distributed and parallel mechanisms or the computational ability of the graphics processing unit (GPU). Parallel solutions should be implemented for incremental learning versions as well because their running time is still much slower than other popular machine learning algorithms such as Naive Bayes and decision trees. Another solution is to apply an approximate nearest neighbor graph to the agglomerative learning algorithm to rapidly find the candidate hyperboxes for aggregation and reduce the number of similarity value computations. Regarding the accuracy, we can consider the multiple values of similarity threshold in the aggregation process rather than only one value as the existing agglomerative learning algorithms.

We can also see that the data presentation orders influence the incremental learning algorithms of the fuzzy min-max neural networks. Therefore, several optimization solutions can be implemented to tackle this problem.

\section{Conclusion and Future work} \label{conclu}
This paper assessed the advantages and drawbacks of the GFMM neural network through empirical results in many benchmark datasets. The impact of setting parameters on the classification problems was also presented. Experimental results indicated the competitive performance of the GFMM neural network compared to other fuzzy min-max systems as well as popular machine learning algorithms using the best parameter settings for each algorithm. Nevertheless, the training time of the GFMM network is a factor preventing the applicability of this type of neural network for the massive datasets in real-world applications.

In future work, we intend to build a novel mechanism to execute the GFMM in parallel for handling massive data. The drawbacks concerning the training time in the agglomerative algorithm will also be enhanced so that we can take advantage of the efficiency of this algorithm for the classification problems in big data. Another potential research direction is the combination of many general fuzzy min-max neural networks at the model level \cite{Gabrys02b}, in which base learners are executed on different clusters in parallel. The automatic manner will be deployed to optimize the hyper-parameters and parameters of learning algorithms aiming at minimizing the generalization errors for each dataset.



\section*{Acknowledgment}
Thanh Tung Khuat would like to acknowledge the FEIT-UTS for awarding him a Ph.D. scholarship.

\bibliographystyle{IEEEtran}
\bibliography{myref}
\end{document}